\documentclass[11pt,a4paper]{article}

\usepackage[T1]{fontenc}
\usepackage[utf8]{inputenc}
\usepackage[margin=1in]{geometry}
\usepackage{amsmath,amssymb,amsthm,mathtools}
\usepackage{bm}
\usepackage{authblk}
\usepackage{graphicx}
\usepackage{float}
\usepackage{booktabs}
\usepackage{multirow}
\usepackage{array}
\usepackage{algorithm}
\usepackage{algpseudocode}
\usepackage[font=small,labelfont=bf]{caption}
\usepackage{xcolor}
\usepackage[numbers,sort&compress]{natbib}
\usepackage[colorlinks=true,linkcolor=blue!60!black,citecolor=blue!60!black,urlcolor=blue!60!black]{hyperref}

\newtheorem{proposition}{Proposition}
\theoremstyle{remark}
\newtheorem{remark}{Remark}

\newcommand{\R}{\mathbb{R}}
\providecommand{\C}{}\renewcommand{\C}{\mathbb{C}}
\providecommand{\B}{}\renewcommand{\B}[1]{\textbf{#1}}
\newcommand{\pmm}{\ensuremath{\,\pm\,}}
\newcommand{\one}{\mathbf{1}}
\newcommand{\Qstat}{Q_{\mathrm{stat}}}
\newcommand{\Rt}{R_{\theta}}
\newcommand{\At}{\widetilde{A}_{\theta}}
\newcommand{\Gt}{\widetilde{G}}
\newcommand{\Pis}{\Pi_{s}}
\newcommand{\knn}{k_{\mathrm{nn}}}
\newcommand{\Favg}{F_{\mathrm{avg}}}
\newcommand{\SA}{\widetilde{S}_{A}}
\newcommand{\sqpi}{\sqrt{\pi}}
\newcommand{\ket}[1]{|#1\rangle}
\newcommand{\bra}[1]{\langle #1|}
\newcommand{\diag}{\operatorname{diag}}
\newcommand{\Tr}{\operatorname{Tr}}
\newcommand{\softmax}{\operatorname{softmax}}
\newcommand{\Acc}{\mathrm{Acc}}
\newcommand{\CNOT}{\mathrm{CNOT}}
\DeclareMathOperator*{\argmax}{arg\,max}
\DeclareMathOperator*{\argmin}{arg\,min}
\algrenewcommand\algorithmicrequire{\textbf{Require:}}
\algrenewcommand\algorithmicensure{\textbf{Ensure:}}

\newcommand{\safefig}[1]{%
\IfFileExists{#1}{\includegraphics[width=\textwidth]{#1}}%
{\fbox{\parbox{0.9\textwidth}{\centering\vspace{2em}Missing figure: \texttt{\detokenize{#1}}\vspace{2em}}}}}

\title{\textbf{Quantum-Informed Regularized Poisson Learning for Graph-Based Semi-Supervised Classification}}

\author[1,6,7]{Hamed Gholipour\thanks{Corresponding author. \texttt{hamed.gholipour@ubi.pt}, \texttt{Hamed.Gholipour@uantwerpen.be}}}
\author[2]{Farid Bozorgnia\thanks{\texttt{f.bozorgnia@newuu.uz}}}
\author[3,4]{Hamzeh Mohammadigheymasi\thanks{\texttt{hmohammadigheymasi@fas.harvard.edu}}}
\author[1]{Jo\~ao Neves\thanks{\texttt{jcneves@ubi.pt}}}
\author[6,7]{Moharram Challenger\thanks{\texttt{moharram.challenger@uantwerpen.be}}}

\affil[1]{Department of Computer Science, University of Beira Interior, R. Marqu\^es de \'Avila e Bolama, Covilh\~a 6201-001, Portugal}
\affil[2]{Department of Mathematics, New Uzbekistan University, Tashkent, Uzbekistan}
\affil[3]{Faculty of Arts and Sciences, Harvard University, 20 Oxford Street, Cambridge, MA 02138, USA}
\affil[4]{Atmosphere and Ocean Research Institute, The University of Tokyo, 5-1-5 Kashiwanoha, Kashiwa, Chiba 277-8564, Japan}
\affil[5]{Instituto de Telecomunica\c{c}\~oes, University of Beira Interior, R. Marqu\^es de \'Avila e Bolama, Covilh\~a 6201-001, Portugal}
\affil[6]{Department of Computer Science, University of Antwerp, Antwerp 2020, Belgium}
\affil[7]{AnSyMo/Cosys Corelab, Flanders Make Strategic Research Center, Leuven 3001, Belgium}

\date{}

\begin{document}
\maketitle

\begin{abstract}
We introduce IPQSSL, a quantum-informed regularized Poisson framework for graph-based semi-supervised classification under scarce labels. Local graph-feature contexts are QR-factorized, lifted to unitary encodings, and processed by a shared variational circuit whose measurements yield node-wise diagonal weights that stabilize a symmetrized Poisson system. We establish positive definiteness of the symmetrized, regularized operator under explicit spectral conditions and derive a Hermitian, block-encoding-compatible formulation that separates the classical NISQ-oriented implementation from a future fault-tolerant pathway. Across seven benchmarks spanning tabular data and a citation network, the effect of regularization is dataset- and budget-dependent: IPQSSL improves over the Poisson baselines on Iris and Wine at all but the single-label budget---reaching 0.946 accuracy on Iris at ten labels per class---and at every budget on German Credit, where, however, all methods remain below the majority-class rate. Original Poisson remains preferable on Breast Cancer, Heart Disease, Optdigits, and Cora, with the deficit narrowing as the label budget grows on the two largest graphs. Ablations attribute the gains to diagonal stabilization rather than variational training. The result is a resource-transparent bridge between Poisson graph learning and quantum linear algebra, with an explicit empirical delineation of when diagonal regularization helps.
\end{abstract}


\section{Introduction}
\label{sec:intro}

Semi-supervised learning (SSL) addresses the setting in which only a small subset of samples is labeled, while a substantially larger set of unlabeled data is available. This setting is common in scientific, industrial, biomedical, and financial applications, where annotation is costly, time-consuming, or requires expert knowledge. The central objective of SSL is therefore to exploit the structure of the unlabeled data to propagate information from the labeled nodes and improve predictive performance under limited supervision~\cite{zhu2003semi}.

Graph-based semi-supervised learning (GSSL) provides a natural framework for this problem. A dataset is represented as a weighted graph, where nodes correspond to samples and edges encode pairwise similarity. Under the manifold assumption, the graph approximates the data geometry discretely, allowing labels to propagate according to local and global connectivity~\cite{zhu2003semi,calder2020poisson}.

However, graph-based propagation methods remain sensitive to the quality of the constructed graph and to the amount of labeled data. Harmonic or Laplacian-based methods may become unstable when the labeled set is extremely small, because label information is imposed mainly through boundary-type constraints. Nonlinear extensions, including graph $p$-Laplacian formulations and their large-$p$ limits, can improve robustness in certain regimes but introduce additional computational and analytical complexity~\cite{kyng2015lipschitz,bozorgnia2026improved}. By contrast, Poisson learning introduces a source-term formulation and has emerged as an effective alternative in very-low-label regimes, often yielding more stable propagation behaviour and enhanced empirical performance~\cite{calder2020poisson}; see also~\cite{bozorgnia2026improved} for an improvement scheme on Poisson learning. This makes Poisson learning a natural candidate for graph-based semi-supervised classification when supervision is scarce and the graph geometry remains informative.

In parallel, advances in quantum computing have stimulated growing interest in machine learning with structured quantum representations. Quantum graph learning aims to encode graph objects, graph operators, or graph-derived features into quantum states and quantum circuits, using superposition, interference, and entanglement as representational resources~\cite{verdon2019qgnn,yu2023qgl,beer2023qml}. This literature includes quantum graph neural networks, equivariant quantum circuits for weighted graphs, and hybrid quantum-classical models designed for noisy intermediate-scale quantum devices (NISQ)~\cite{skolik2023equivariant,ceschini2024graphs}.

A fundamental obstacle is that classical graph-propagation operators are generally non-unitary, whereas closed-system quantum evolution must be represented by unitary transformations. This mismatch is particularly evident in Laplacian and Poisson propagation, where the learning rule is naturally expressed as a linear system or an affine fixed-point iteration rather than a unitary evolution. In addition, exact state preparation for dense graph-derived matrices can be prohibitively expensive, so any practical hybrid framework must balance representational richness against circuit depth, gate count, and noise sensitivity on near-term hardware~\cite{schuld2019feature,schuld2021book}. Block encoding, linear-combination-of-unitaries constructions, and Quantum Singular Value Transformation (QSVT) provide principled ways to embed non-unitary matrices into larger unitary operators~\cite{gilyen2019qsvt,childs2012lcu,harrow2009hhl}. However, such constructions rely on explicit assumptions about access, state-preparation routines, conditioning, and readout costs. Consequently, a responsible quantum-informed formulation must specify which parts of the pipeline are implemented classically, which quantities are generated by quantum circuits, and which components are reserved for a future fault-tolerant implementation pathway.

Motivated by these considerations, this paper develops a quantum-informed regularized Poisson framework for graph-based semi-supervised classification under limited labeled data. The primary model, called Improved Poisson Quantum Semi-Supervised Learning (IPQSSL), combines classical Poisson label propagation with QR-decomposition-based quantum-compatible encodings of local graph-feature contexts. For each node, a local context matrix is constructed from the node and its graph neighbours, factorized through QR decomposition, embedded into a unitary-compatible representation, and processed by a parameterized quantum circuit. Measurements of this circuit generate observable-dependent diagonal weights which act as node-wise regularization terms in the Poisson system.

The present implementation should be understood as quantum-informed rather than as an end-to-end quantum algorithm. Graph construction, QR preprocessing, and the solution of the Poisson linear system are performed classically. The quantum component contributes through the circuit-generated diagonal regularizer and through circuit-level diagnostics such as entanglement entropy and randomised-benchmarking-inspired fidelity. In addition, we derive a symmetrized, Hermitian, block-encoding-compatible formulation of the regularized Poisson operator, identifying how the same operator could be embedded into a future quantum linear-system workflow under suitable sparse-access, state-preparation, conditioning, and readout assumptions.

This work builds on our earlier Laplacian Quantum Semi-Supervised Learning (LQSSL) model~\cite{gholipour2025lqssl}, published in \emph{Quantum Information Processing}, which uses the normalized graph Laplacian as the sole propagation operator and embeds graph-derived features into a parameterized quantum circuit without an explicit node-wise regularization mechanism. IPQSSL departs from LQSSL along two axes. First, it replaces Laplacian propagation with an improved Poisson source-term formulation, motivated by the well-documented instability of harmonic extension under very low label rates~\cite{calder2020poisson}. Second, it introduces the QR-based quantum-compatible encoding as a source of adaptive node-wise diagonal regularization weights for the Poisson system. Because any diagonal regularization could in principle improve the conditioning of the propagation operator regardless of its origin, the diagonal-regularisation ablation of Section~\ref{sec:ablation} is used to separate the contribution of diagonal stabilization itself from that of the trained quantum component, rather than attributing observed gains to the quantum circuit by default.

The main contributions of this work are as follows.
\begin{enumerate}
\item \textbf{QR-based quantum-compatible node encoding.} We introduce a QR-decomposition-based mechanism that maps local graph-feature context matrices into unitary-compatible node representations. In the present implementation, this representation does not replace classical graph propagation. Instead, it produces observation-dependent diagonal weights that regularize the Poisson system node-wise. Whether these weights add value beyond generic bounded diagonals is tested, not assumed: the ablation of Section~\ref{sec:ablation} shows that, under the present protocol, they are competitive with but not superior to simpler constructions.
\item \textbf{Block-encoding-compatible formulation of the regularized Poisson operator.} We derive the symmetrized Hermitian operator and show that it admits a block-encoding-compatible representation under standard sparse-access, degree-access, stationary-state-preparation, and diagonal-access assumptions. This analysis clarifies the distinction between the current classical linear solve and a possible future fault-tolerant quantum linear-system implementation.
\item \textbf{Circuit-level diagnostics.} We report entanglement-entropy- and randomised-benchmarking-inspired diagnostics to analyze how circuit depth and width affect the QR-based quantum-compatible representation. These diagnostics are interpreted as circuit-level evidence of expressivity and noise sensitivity, not as a substitute for predictive performance or as evidence of quantum advantage.
\end{enumerate}

The results show that regularized Poisson propagation is effective when the similarity graph adequately captures the data geometry, although performance remains dataset-dependent. The main theoretical contribution is a symmetrized Hermitian formulation of the regularized Poisson operator that supports spectral analysis and is compatible, at the operator level, with block-encoding-based quantum linear algebra methods.

The remainder of the paper is organized as follows. Section~\ref{sec:background} reviews graph-based semi-supervised learning, Poisson learning, quantum graph learning, and block-encoding concepts. Section~\ref{sec:method} develops the proposed framework, including the improved Poisson propagation rule, the QR-based quantum-compatible encoding, the symmetrized regularized Poisson operator, and the IPQSSL algorithm. Section~\ref{sec:protocol} describes the experimental protocol, datasets, baselines, ablations, and evaluation metrics. Section~\ref{sec:results} reports the predictive and ablation results, while Section~\ref{sec:circuit} presents circuit-level entanglement and randomized-benchmarking-inspired diagnostics. Section~\ref{sec:discussion} discusses interpretation and applicability, and Section~\ref{sec:conclusion} concludes with future research directions.

\section{Background}
\label{sec:background}

This section establishes the conceptual foundations required for the proposed IPQSSL framework. We first review graph-based semi-supervised learning and the low-label motivation for Poisson propagation, then summarize hybrid quantum graph learning and the representation of graph-derived operators in a quantum-compatible form. The section concludes by connecting these ingredients to the design requirements that motivate the framework developed in Section~\ref{sec:method}.

\subsection{Graph-based semi-supervised learning}
\label{sec:gssl}

This subsection fixes the graph notation used throughout the paper and recalls the propagation setting in which Poisson learning becomes particularly relevant under limited supervision.

Graph-based semi-supervised learning (GSSL) assumes that a dataset can be represented by a weighted graph $G=(V,E,W)$, where each node corresponds to a sample, and each weighted edge encodes a notion of similarity between two samples. Let $n=|V|$, let $W\in\R^{n\times n}$ denote the similarity matrix, and let $D=\diag(d_1,\dots,d_n)$ be the degree matrix, with $d_i=\sum_j W_{ij}$.

A central advantage of GSSL is that it exploits relational structure rather than relying only on pointwise classification. This is particularly useful when the data are assumed to lie near a lower-dimensional manifold so that nearby points in the graph are likely to share labels. Classical propagation methods, therefore, use graph operators derived from $W$ and $D$ extending label information across the graph while respecting both local neighborhood structure and global connectivity~\cite{zhu2003semi,calder2020poisson}.

Two important families of graph-based propagation methods are Laplacian learning and Poisson learning. In Laplacian-based methods, label propagation is typically formulated through a graph-harmonic extension problem involving the graph Laplacian
\[
L = D - W .
\]
This approach provides a natural and mathematically elegant framework for manifold-based classification, but its performance may deteriorate when the labeled set is very small or when the graph is only weakly informative. Nonlinear extensions, including $p$-Laplacian formulations, can improve robustness in certain regimes, although at the cost of additional computational and analytical complexity~\cite{elalaoui2016asymptotic}. By contrast, Poisson learning has emerged as an effective alternative in the low-label regime by reformulating label propagation through a graph Poisson equation, often leading to improved empirical stability when supervision is extremely scarce~\cite{calder2020poisson}.

\subsection{Quantum graph learning and hybrid quantum-classical models}

Recent progress in quantum computing has motivated growing interest in quantum graph learning (QGL), which seeks to encode graph structure, graph operators, or graph-derived features into quantum states and quantum circuits. Broadly speaking, this literature includes quantum algorithms operating on graph-structured inputs, quantum graph representations, and quantum graph neural networks (QGNNs). In the Noisy Intermediate-Scale Quantum (NISQ) setting, the most practically relevant approaches are hybrid quantum-classical models that combine classical preprocessing and optimization with parameterized quantum circuits~\cite{ceschini2024graphs}.

From a machine-learning perspective, QGL is motivated by the possibility that quantum state spaces may provide expressive representations for structured data. Parameterized quantum circuits can, in principle, encode graph information into amplitudes, phases, and entanglement patterns, thereby offering a richer representational layer than purely classical feature maps.

Current quantum hardware imposes substantial limitations. NISQ devices provide only a modest number of qubits, are affected by gate noise and decoherence, and make deep circuit constructions difficult to execute reliably~\cite{preskill2018nisq}. These constraints are especially relevant for graph-learning tasks, since realistic graphs may be high-dimensional, dense, or irregular. For this reason, hybrid approaches are particularly important: the classical component can handle graph construction, matrix preprocessing, and outer-loop optimization, while the quantum component encodes and analyzes structured transformations within a circuit-based model~\cite{bharti2022nisq}.

\subsection{Quantum representation of graph-derived operators}

A central challenge in hybrid quantum formulations of graph-based learning is that the classical propagation operators of interest are generally non-unitary, whereas quantum evolution must be implemented through unitary transformations. This issue arises immediately for Laplacian- and Poisson-type updates, which are naturally expressed in terms of linear or affine operators on label matrices rather than as unitary quantum gates~\cite{gilyen2019qsvt}.

A standard first step in quantum graph learning is therefore to encode graph-derived information into quantum states. In this work, the relevant objects are constructed from the similarity matrix and related graph operators, then embedded into parameterized quantum circuits through a QR-decomposition-based procedure. At a high level, QR decomposition provides an orthonormal factor that can be interpreted as a unitary-compatible transformation for state preparation. This makes it a convenient bridge between classical graph matrices and circuit-level quantum representations~\cite{schuld2021book,beer2023qml}.

More generally, amplitude encoding provides a natural way to represent graph-derived vectors within quantum states. If $\beta=(\beta_1,\dots,\beta_n)$ is a normalized vector derived from graph data, then its quantum representation takes the form
\[
\ket{\psi_\beta} = \sum_{i=1}^{n} \beta_i \ket{i},
\]
so that the amplitudes encode node- or graph-level information. This representation is conceptually attractive because it links graph structure to the geometry of quantum states, although its practical usefulness depends strongly on the cost of state preparation and on the choice of encoding map.

The second challenge is implementing non-unitary updates. Tools from quantum linear algebra, particularly block encoding and linear-combination constructions, provide a principled route for embedding non-unitary matrices into larger unitary operators. These techniques are therefore relevant to graph-based semi-supervised learning, where the iterative propagation rule must be translated into a quantum-compatible form. In the present work, they serve as the conceptual foundation for the hybrid treatment of the propagation operator developed in the next section~\cite{childs2012lcu}.

\subsection{Motivation for the present framework}

The preceding discussion highlights three requirements for a useful hybrid quantum approach to graph-based semi-supervised learning. First, the method must preserve the geometric role of the similarity graph, since label propagation remains fundamentally graph-driven. Second, it must provide a practical mechanism for embedding graph-derived structure into a unitary quantum representation. Third, it must address the mismatch between classical non-unitary propagation rules and quantum circuit constraints without obscuring the cost of state preparation or the limitations of near-term hardware.

These requirements motivate the framework developed in this paper. We combine improved Poisson-based propagation with QR-decomposition-based quantum encoding and a block-encoding-inspired treatment of the resulting non-unitary update. This formulation is intended not as a claim of unconditional quantum advantage, but as a hybrid framework for studying how graph propagation, quantum representation, and hardware-sensitive circuit behaviour interact under limited supervision.

\section{Enhancement of Quantum Graph-Based Semi-Supervised Learning}
\label{sec:method}

This section develops the proposed framework through four connected components. Section~\ref{sec:ipl} reviews the Improved Poisson Learning (IPL) propagation rule~\cite{bozorgnia2026improved}, which forms the classical backbone of the model. Section~\ref{sec:qr} introduces the QR-based quantum-compatible encoding that generates node-wise regularization weights, while Section~\ref{sec:block} derives the Hermitian block-encoding-compatible form of the resulting regularized Poisson operator. Finally, Section~\ref{sec:ipqssl} combines these components into the complete Improved Poisson Quantum Semi-Supervised Learning (IPQSSL) procedure summarized in Algorithm~\ref{alg:ipqssl}.

Relative to our earlier LQSSL model~\cite{gholipour2025lqssl}, which propagates labels through the normalized graph Laplacian and contains no node-wise regularization mechanism, IPQSSL changes both the propagation paradigm and the role of the quantum circuit: label propagation is carried out through the improved Poisson formulation of Equation~\eqref{eq:ipl_iter}, and the QR-based quantum encoding is used to generate adaptive diagonal weights $\Rt$ that regularize the Poisson system. The shift from Laplacian to Poisson propagation is motivated by the well-documented instability of harmonic extension under very low label rates~\cite{calder2020poisson}, which the source-term formulation of Poisson learning is designed to address. Whether the quantum-derived diagonal weights add value beyond generic diagonal stabilization is treated as an empirical question and is assessed through the ablation of Section~\ref{sec:ablation}.

\subsection{Review of Improved Poisson Learning for Graph-Based Semi-Supervised Learning}
\label{sec:ipl}

In this subsection, we briefly review the Improved Poisson Learning (IPL) scheme introduced in~\cite{bozorgnia2026improved}, which forms the classical propagation mechanism underlying the quantum-informed construction developed in Section~\ref{sec:qr}. Poisson learning is particularly attractive in the low-label regime because, rather than enforcing the available labels through pointwise boundary-type constraints, it interprets them as sources and sinks of a graph Poisson equation. This formulation mitigates the degeneracy that may arise in harmonic or Laplacian-based label propagation when the fraction of labeled nodes becomes very small.

Let $W\in\R^{n\times n}$ be the symmetric similarity matrix introduced in Section~\ref{sec:gssl}, let
\[
D=\diag(d_1,\dots,d_n), \qquad d_i=\sum_{j=1}^{n} W_{ij},
\]
and define the random-walk transition matrix
\[
P = D^{-1}W .
\]
Since $P\one=\one$, the matrix $P$ is row stochastic and therefore has the eigenvalue $1$.

Let $\mathcal{L}\subset\{1,\dots,n\}$ denote the set of labeled nodes, with $|\mathcal{L}|=m_{\ell}$, and let $Y\in\R^{n\times C}$ denote the label matrix. Writing $y_i^{\top}$ for the $i$-th labeled row, define the mean label vector
\[
\bar{y} = \frac{1}{m_{\ell}}\sum_{i\in\mathcal{L}} y_i \in \R^{C}.
\]
The centered Poisson source matrix $B\in\R^{n\times C}$ is then defined by
\begin{equation}
B_{i:} =
\begin{cases}
y_i^{\top}-\bar{y}^{\top}, & i\in\mathcal{L},\\[2pt]
0, & i\notin\mathcal{L}.
\end{cases}
\label{eq:source}
\end{equation}
Equivalently, after ordering the labeled nodes first,
\[
B = \begin{bmatrix} Y_{\ell}-\one_{m_{\ell}}\bar{y}^{\top} \\ 0 \end{bmatrix}.
\]
By construction,
\begin{equation}
\one^{\top}B = 0,
\label{eq:centered}
\end{equation}
so the forcing term contains no constant class-wise component.

The standard iterative realization of graph Poisson learning, with iteration index $t$, can be written as
\begin{equation}
U^{(t+1)} = P\,U^{(t)} + D^{-1}B .
\label{eq:poisson_iter}
\end{equation}
Indeed, this follows from the usual update
\[
U^{(t+1)} = U^{(t)} + D^{-1}\bigl(B-(D-W)U^{(t)}\bigr),
\]
since $P=D^{-1}W$. The difficulty is that $P$ possesses the stationary eigenvalue $1$. Consequently, the homogeneous propagation does not define a strict contraction, and the stationary component of the random walk may dominate the long-time dynamics.

For a connected undirected graph, the stationary distribution of $P$ is
\[
\pi = \Bigl(\frac{d_1}{d},\dots,\frac{d_n}{d}\Bigr)^{\top}, \qquad d=\sum_{i=1}^{n} d_i,
\]
and satisfies
\[
\pi^{\top}P=\pi^{\top}.
\]
We therefore introduce the rank-one stationary projection
\begin{equation}
\Qstat = \one\pi^{\top}.
\label{eq:qstat}
\end{equation}
It satisfies
\[
\Qstat^{2}=\Qstat, \qquad P\Qstat=\Qstat P=\Qstat .
\]
Thus $\Qstat$ isolates the invariant stationary mode of the random-walk dynamics. Subtracting this component suppresses the long-time drift toward the stationary distribution and constitutes the main modification introduced by IPL.

The generalized IPL iteration used here is
\begin{equation}
U^{(t+1)} = \bigl(P-\alpha_1\Qstat+\alpha_2 I\bigr)U^{(t)} + \alpha_3 D^{-1}B,
\label{eq:ipl_iter}
\end{equation}
with initialization
\begin{equation}
U^{(0)} = D^{-1}Y,
\label{eq:ipl_init}
\end{equation}
where $\alpha_1,\alpha_2,\alpha_3\in\R$ are tunable parameters. The term $-\alpha_1\Qstat$ controls the stationary mode, $\alpha_2 I$ introduces a self-retention contribution, and $\alpha_3$ controls the strength of the Poisson source injection.

Introducing the propagation operator
\begin{equation}
M := P-\alpha_1\Qstat+\alpha_2 I,
\label{eq:M}
\end{equation}
the iteration becomes
\begin{equation}
U^{(t+1)} = M U^{(t)} + \alpha_3 D^{-1}B .
\label{eq:ipl_M}
\end{equation}
This decomposition is particularly useful in the present work because it separates the graph propagation operator $M$ from the external source term. The former will subsequently constitute the principal non-unitary operator associated with the quantum-informed formulation.

To clarify the convergence mechanism, note first that
\[
M\one = (1-\alpha_1+\alpha_2)\one .
\]
Hence the stationary eigenvalue $1$ of $P$ is shifted to
\[
1-\alpha_1+\alpha_2 .
\]
Moreover, the subspace
\[
H_0 = \bigl\{v\in\R^{n} : \pi^{\top}v=0\bigr\}
\]
is invariant under $P$. Since $\Qstat v=0$ for $v\in H_0$, the action of $M$ on this subspace reduces to
\[
Mv=(P+\alpha_2 I)v .
\]
Because $W$ is symmetric, $P=D^{-1}W$ is similar to the symmetric matrix $D^{-1/2}WD^{-1/2}$, and its eigenvalues are real. If
\[
1=\lambda_1(P),\ \lambda_2(P),\ \dots,\ \lambda_n(P)
\]
denote the eigenvalues of $P$, then
\begin{equation}
\sigma(M) = \bigl\{1-\alpha_1+\alpha_2,\ \lambda_2(P)+\alpha_2,\ \dots,\ \lambda_n(P)+\alpha_2\bigr\}.
\label{eq:specM}
\end{equation}
Consequently, a sufficient and, for this linear iteration, natural convergence condition is
\begin{equation}
\rho(M) = \max\Bigl\{|1-\alpha_1+\alpha_2|,\ \max_{2\le j\le n}|\lambda_j(P)+\alpha_2|\Bigr\} < 1 .
\label{eq:rhoM}
\end{equation}
Under this condition,
\[
M^{t}\longrightarrow 0 \quad\text{as } t\to\infty,
\]
and the IPL iteration converges for every initialization to the unique fixed point
\begin{equation}
U^{*} = \alpha_3 (I-M)^{-1}D^{-1}B .
\label{eq:fixedpoint}
\end{equation}
Equivalently,
\begin{equation}
U^{*} = \alpha_3\sum_{r=0}^{\infty} M^{r}D^{-1}B .
\label{eq:neumann}
\end{equation}

The centered structure of $B$ is also compatible with the stationary projection. Indeed,
\[
\pi^{\top}D^{-1} = \frac{1}{d}\one^{\top},
\]
and therefore
\begin{equation}
\Qstat D^{-1}B = \one\pi^{\top}D^{-1}B = \frac{1}{d}\one\one^{\top}B = 0,
\label{eq:stat_source}
\end{equation}
where~\eqref{eq:centered} has been used. Thus the injected Poisson source has no component along the stationary random-walk mode. This centering is important in class-imbalanced settings because it prevents the source term itself from reinforcing the degree-weighted stationary bias.

The scalar parameter $\alpha_3$ affects the amplitude of the source-driven steady state but does not alter the spectral radius of $M$. Hence the convergence rate of the homogeneous propagation is determined entirely by the spectral properties of $M$, whereas $\alpha_3$ controls the strength of label injection. In practice, the iteration is terminated when
\begin{equation}
\|U^{(t)}-U^{(t-1)}\|_{F} < \varepsilon,
\label{eq:stop}
\end{equation}
for a prescribed tolerance $\varepsilon>0$. The final class prediction is then obtained row-wise according to
\begin{equation}
\hat{y}_i = \argmax_{1\le c\le C} U^{*}_{ic}.
\label{eq:predict}
\end{equation}

A remark is needed to reconcile this convergence analysis with the implementation used in the remainder of the paper. Condition~\eqref{eq:rhoM} governs only the iterative realization~\eqref{eq:ipl_M}. At the default hyperparameters of Table~\ref{tab:hyper}, the spectral radius exceeds unity on every dataset ($\rho(M)\in[1.038,1.100]$; Table~\ref{tab:spectral}), so the iteration itself would diverge in that regime. All experiments in this paper therefore compute the fixed point by solving the linear system $(I-M)U^{*}=\alpha_3 D^{-1}B$ directly, in the symmetrized form derived in Section~\ref{sec:block}, for which only $1\notin\sigma(M)$ is required rather than $\rho(M)<1$. The stopping rule~\eqref{eq:stop} applies only when the iterative form is used.

From the perspective of the present work, the essential outcome of IPL is therefore not merely a modified label propagation rule, but the explicit non-unitary graph operator $M$ in~\eqref{eq:M}. Its stationary-mode correction, spectral structure, and affine source-driven dynamics provide the classical foundation for the hybrid quantum--classical construction developed next.

\subsection{QR-Based Quantum-Compatible Encoding of Local Graph Contexts}
\label{sec:qr}

Having established the Improved Poisson Learning framework in Section~\ref{sec:ipl}, we now introduce the quantum-informed component of IPQSSL. The purpose of this component is deliberately specific: it does not replace the Poisson propagation step, but instead learns a node-dependent diagonal regularizer from the local graph--feature geometry. The resulting weights modulate the classical propagation operator introduced in the preceding section.

The construction proceeds in four stages. For each node, we first form a local context matrix containing the node and its nearest neighbours. We then extract an orthogonal representation of this local context through a QR factorization and embed the resulting representation into a small qubit register. A single variational circuit, shared across all nodes, maps each encoded state to a bounded scalar score. Finally, these scores are assembled into a diagonal matrix
\[
\Rt = \diag\bigl(r_1(\theta),\dots,r_n(\theta)\bigr),
\]
which enters the regularized Poisson system developed in the next subsection. This separation makes the role of the quantum component explicit: classical graph construction and Poisson propagation are retained, while the circuit provides a trainable rule for assigning node-wise regularization weights from local geometric information.

\paragraph{Local graph contexts and QR representation.}
For each node $i$, let $\mathcal{N}_i$ denote the set of its $\knn$ nearest neighbours in the similarity graph constructed in Section~\ref{sec:graph}. To make the encoding deterministic, the neighbours are ordered by increasing feature-space distance, with the node index used to break ties. If
\[
\mathcal{N}_i=\{j_1,\dots,j_{\knn}\},
\]
the local context matrix associated with node $i$ is defined as
\begin{equation}
A_i =
\begin{bmatrix}
x_i^{\top}\\ x_{j_1}^{\top}\\ \vdots\\ x_{j_{\knn}}^{\top}
\end{bmatrix}
\in\R^{m\times d}, \qquad m=\knn+1,
\label{eq:context}
\end{equation}
where $x_r\in\R^{d}$ denotes the feature vector of node $r$. Thus, it $A_i$ represents the feature geometry of the node $i$ together with its local graph neighborhood. (Here and below, $m$ denotes the number of rows of the local context; it is distinct from the labeled-set size $m_{\ell}$ of Section~\ref{sec:ipl}.)

We compute an economy-size QR factorization
\begin{equation}
A_i = Q_iR_i, \qquad Q_i^{\top}Q_i=I,
\label{eq:qr}
\end{equation}
with the signs chosen so that the relevant diagonal entries $R_i$ are non-negative. This convention removes the otherwise arbitrary sign ambiguity of the QR representation and makes the encoding reproducible.

For clarity, we first describe the generic full-rank case, for which
\[
s=\min(m,d), \qquad Q_i\in\R^{m\times s}, \qquad R_i\in\R^{s\times d}.
\]
If $A_i$ is numerically rank deficient; only the directions retained under the fixed numerical-rank tolerance used in the implementation are included in the subsequent encoding.

The columns of $Q_i$ provide orthonormal directions describing the local context of node $i$, whereas $R_i$ contains the corresponding scale information. We encode only $Q_i$. Consequently, the quantum representation is constructed from normalized local-context directions rather than directly from the raw feature magnitudes. In the generic full-rank setting and for a fixed neighbourhood structure, the chosen QR convention is also insensitive to positive rescaling of individual feature columns. This invariance should be interpreted locally: changing the feature scaling before graph construction may alter pairwise distances, the neighbour sets, the kernel bandwidth, and hence the graph itself.

Two additional properties of the representation are important. First, the encoding is not permutation invariant with respect to the neighbours, since rows of $A_i$ are indexed according to their distance-based ordering. This ordering is intentional because it preserves neighbourhood-rank information, although it introduces a discontinuity when a perturbation exchanges the order of two neighbours. Second, the QR representation depends on the ordering of feature coordinates. Unlike neighbour ordering, this dependence has no geometric interpretation and is inherited from the representation of the input data. We retain these conventions because they provide a deterministic and reproducible node-wise map.

\paragraph{Embedding into a qubit register.}
We next convert the orthogonal local representation into a normalized quantum state. Let
\[
q\ge\lceil\log_2 m\rceil, \qquad N_q=2^{q},
\]
so that the Hilbert-space dimension is at least the number of rows of $Q_i$. Unless stated otherwise, we use the smallest admissible register,
\[
q_{\min}=\lceil\log_2 m\rceil .
\]
If $m<N_q$, $Q_i$ is embedded into the register dimension by zero-padding:
\begin{equation}
\bar{Q}_i = \begin{bmatrix} Q_i\\ 0 \end{bmatrix}\in\R^{N_q\times s}, \qquad \bar{Q}_i^{\top}\bar{Q}_i=I_s .
\label{eq:pad}
\end{equation}
Hence $\bar{Q}_i$ is an isometry from the $s$-dimensional QR-coordinate space into the $N_q$-dimensional qubit Hilbert space.

Define
\[
c=\frac{1}{\sqrt{s}}\one_s .
\]
The QR-encoded state of node $i$ is
\begin{equation}
\ket{\psi_i} = \bar{Q}_i c = \frac{1}{\sqrt{s}}\sum_{a=0}^{s-1}\ket{\bar{q}_{i,a}},
\label{eq:psi}
\end{equation}
where $\bar{q}_{i,a}$ denotes the $a$-th column of $\bar{Q}_i$. Because the columns of $\bar{Q}_i$ are orthonormal,
\[
\langle\psi_i|\psi_i\rangle = c^{\top}\bar{Q}_i^{\top}\bar{Q}_i c = c^{\top}c = 1,
\]
so no additional normalization is required.

For a quantum-circuit interpretation, $\bar{Q}_i$ can be completed deterministically to an orthogonal matrix
\[
\widetilde{Q}_i = [\,\bar{Q}_i\ \ Q_i^{\perp}\,]\in\R^{N_q\times N_q}.
\]
Since a real orthogonal matrix is also unitary, define the reference $q$-qubit state
\begin{equation}
\ket{u} = \frac{1}{\sqrt{s}}\sum_{a=0}^{s-1}\ket{a}.
\label{eq:ref_state}
\end{equation}
Then
\[
\ket{\psi_i} = \widetilde{Q}_i\ket{u}.
\]
The completion $Q_i^{\perp}$ has no effect on the prepared state because $\ket{u}$ has support only on the first $s$ computational basis states. It is therefore required only to provide a mathematically valid unitary interpretation of the data-loading map; it is not constructed in the classical implementation.

\paragraph{Shared variational circuit and node-wise score.}
The encoded state $\ket{\psi_i}$ summarizes the local context of node $i$, but the regularization weight must still be learned from the classification task. To this end, every node is processed by the same parameterized circuit $V(\theta)$:
\begin{equation}
\ket{\phi_i(\theta)} = V(\theta)\ket{\psi_i}.
\label{eq:phi}
\end{equation}
Only the input state varies across nodes; the circuit parameters $\theta$ are shared globally. This parameter sharing keeps the number of trainable parameters independent of the graph size and forces the model to learn a common rule for scoring local neighbourhood geometries rather than a separate parameter for every node.

We use $L_q$ hardware-efficient layers of single-qubit rotations followed by a ring of CNOT entanglers:
\begin{equation}
V_{\ell}(\theta_{\ell}) = U_{\mathrm{ent}}\,U_{\mathrm{rot}}(\theta_{\ell}),
\label{eq:layer}
\end{equation}
where
\[
U_{\mathrm{rot}}(\theta_{\ell}) = \bigotimes_{a=0}^{q-1} R_z(\theta_{\ell,a,3})\,R_y(\theta_{\ell,a,2})\,R_x(\theta_{\ell,a,1}),
\]
and
\[
U_{\mathrm{ent}} = \prod_{a=0}^{q-1}\CNOT_{a\to(a+1)\bmod q}.
\]
The complete circuit is
\begin{equation}
V(\theta) = V_{L_q}(\theta_{L_q})\cdots V_1(\theta_1),
\label{eq:circuit}
\end{equation}
with
\[
N_{\theta}=3qL_q
\]
trainable parameters. The $R_zR_yR_x$ block gives a general single-qubit rotation up to global phase, while repeated nearest-neighbour entangling layers allow correlations to propagate throughout the register without requiring an all-to-all entangling topology.

The circuit is read out using the fixed observable
\[
Z_0 = Z\otimes I^{\otimes(q-1)}.
\]
For each node, we define
\begin{equation}
r_i(\theta) = \frac{1+\langle\phi_i(\theta)|Z_0|\phi_i(\theta)\rangle}{2}.
\label{eq:score}
\end{equation}
Since the expectation value of $Z_0$ belongs to $[-1,1]$,
\begin{equation}
0\le r_i(\theta)\le 1 .
\label{eq:score_bound}
\end{equation}
Thus, the circuit directly produces a bounded and non-negative scalar weight, without requiring clipping or post-processing.

Collecting the node-wise scores gives
\begin{equation}
\Rt = \diag\bigl(r_1(\theta),\dots,r_n(\theta)\bigr).
\label{eq:Rtheta}
\end{equation}
This matrix is the only object passed from the quantum-informed component to the subsequent Poisson propagation stage.

\paragraph{Classical interpretation of the circuit score.}
It is useful to characterize explicitly what function the circuit represents. This both clarifies the contribution of the quantum component and prevents a computational interpretation that is stronger than what the present architecture supports.

Let $\psi_i\in\C^{N_q}$ denote the amplitude vector of $\ket{\psi_i}$, and define the parameter-dependent observable
\begin{equation}
O_{\theta} = V(\theta)^{\dagger}Z_0V(\theta).
\label{eq:Otheta}
\end{equation}
Equation~\eqref{eq:score} can then be written as
\begin{equation}
r_i(\theta) = \frac{1+\psi_i^{\dagger}O_{\theta}\psi_i}{2}.
\label{eq:score_quad}
\end{equation}
Hence, for fixed $\theta$, the circuit evaluates a bounded quadratic form in the QR-encoded state.

Because $O_{\theta}$ is obtained by unitary conjugation of $Z_0$, it is Hermitian and satisfies
\[
O_{\theta}^{2}=I, \qquad \sigma(O_{\theta})\subseteq\{-1,+1\}.
\]
However, $O_{\theta}$ is not an arbitrary Hermitian involution. It belongs to the restricted family
\[
\mathcal{O} = \bigl\{V(\theta)^{\dagger}Z_0V(\theta) : \theta\in\Theta\bigr\},
\]
whose accessible geometry is determined by the chosen ansatz and its $N_{\theta}=3qL_q$ parameters. The circuit therefore realizes a structured, low-parameter family of quadratic forms rather than an unrestricted $N_q\times N_q$ Hermitian model.

Moreover, because the QR-encoded amplitudes used here are real,
\[
\psi_i^{\dagger}O_{\theta}\psi_i = \psi_i^{\top}\operatorname{Re}(O_{\theta})\psi_i,
\]
so only the real symmetric component of the effective observable contributes to the score on the present data representation.

This characterization is important for interpreting the method. At the small register sizes considered in this work, the same quadratic form can be evaluated efficiently on a classical computer. Accordingly, the role of the circuit is not to provide a demonstrated computational speedup. Rather, it provides a hardware-compatible and structurally constrained parameterization of the mapping
\[
\ket{\psi_i}\longmapsto r_i(\theta),
\]
with unitarity and circuit architecture imposing an inductive bias on the admissible score functions. This is the precise sense in which the method is described as quantum-informed.

The same observation also identifies natural matched-capacity classical controls. In particular, a directly parameterized quadratic form acting on $\psi_i$ isolates the effect of the circuit constraint, while a small feed-forward network of comparable parameter count provides a generic nonlinear alternative. These comparisons help distinguish the contribution of the QR representation from that of the variational parameterization.

\paragraph{Supervised learning of the diagonal regularizer.}
The circuit parameters are not chosen independently of the downstream task. Instead, $\Rt$ is trained through the classification performance of the regularized Poisson system.

Let
\[
F^{\star}_{\theta}\in\R^{n\times C}
\]
denote the label-score matrix obtained from the Poisson system corresponding to $\Rt$, and define
\[
p_i(\theta) = \softmax\bigl(F^{\star}_{\theta}(i,:)\bigr).
\]
The circuit parameters are obtained by minimizing the supervised cross-entropy over the labeled set:
\begin{equation}
\theta^{\star} = \argmin_{\theta}\ \mathcal{L}_{\mathrm{sup}}(\theta),
\label{eq:theta_star}
\end{equation}
where
\begin{equation}
\mathcal{L}_{\mathrm{sup}}(\theta) = -\frac{1}{|\mathcal{L}|}\sum_{i\in\mathcal{L}}\sum_{c=1}^{C} Y_{ic}\log p_{ic}(\theta) + \frac{\lambda_{\theta}}{2}\|\theta\|_2^{2},
\label{eq:loss}
\end{equation}
and $\lambda_{\theta}\ge 0$ is an optional weight-decay parameter.

The optimization is naturally bilevel: $\theta$ influences the loss through the solution $F^{\star}_{\theta}$ of the inner Poisson system. Consequently, each objective evaluation requires both evaluating the node-wise circuit scores and solving the corresponding linear system.

One possibility would be to differentiate explicitly through the Poisson solver, either by unrolling its iterations or by implicit differentiation. Parameter-shift differentiation would instead require a number of circuit--solver evaluations proportional to $N_{\theta}$ at every optimization step. In the present implementation, we use simultaneous perturbation stochastic approximation (SPSA), which estimates the update direction using two perturbed objective evaluations per optimization iteration, independently of $N_{\theta}$. This is particularly convenient for the present hybrid architecture because the Poisson solver can be treated as a black box and the same training procedure remains applicable when expectation values are estimated from finite-shot circuit measurements. The SPSA settings (number of iterations, gain sequences, initialization $\theta_0$, and stopping criterion) and the value of $\lambda_{\theta}$ are listed in Table~\ref{tab:hyper}; in all experiments reported here the expectation values are evaluated exactly by statevector simulation.

The resulting trained regularizer is denoted by
\begin{equation}
R_{\theta^{\star}} = \diag\bigl(r_1(\theta^{\star}),\dots,r_n(\theta^{\star})\bigr).
\label{eq:Rtheta_star}
\end{equation}

\paragraph{Computational scope and implementation cost.}
The construction above should be interpreted as a hybrid, quantum-compatible regularization mechanism rather than as an end-to-end quantum algorithm. In the implementation considered here, we perform graph construction, formation of the local matrices $A_i$, QR factorization, and the Poisson solve classically. The quantum component is restricted to evaluating the expectation values in Eq.~\eqref{eq:score}, from which we construct $R_{\theta^{\star}}$.

For each node, the QR factorization of
\[
A_i\in\R^{(\knn+1)\times d}
\]
requires
\[
O\bigl((\knn+1)\,d\,s\bigr)
\]
operations, where $s=\min(\knn+1,d)$ in the generic full-rank case. For fixed $\knn$ and feature dimension $d$, the aggregate QR preprocessing cost therefore scales linearly with the number of nodes. Zero-padding from $Q_i$ to $\bar{Q}_i$ is negligible, and the orthogonal completion $\widetilde{Q}_i$ is never formed numerically.

The main quantum resource issue is state preparation. Preparing a generic $q$-qubit amplitude state may require $O(2^{q})$ two-qubit gates in the worst case~\cite{shende2006synthesis,reck1994unitary}. Synthesizing the complete unitary $\widetilde{Q}_i$ would be substantially more expensive and is unnecessary because the algorithm requires only the state $\ket{\psi_i}$, not the complete basis transformation. Accordingly, the QR encoding is practically relevant here because the local registers are deliberately small.

More generally, a future fault-tolerant implementation would have to account for the costs of graph construction, local QR preprocessing, state preparation, expectation-value estimation, optimization, preparation of the Poisson right-hand side, and readout. Any claim involving block encoding, quantum singular-value transformation, or a quantum linear-system algorithm would therefore require these costs to be included explicitly. We consequently make no claim of an end-to-end quantum computational advantage in the present work.

\paragraph{Interpretation and connection to Poisson propagation.}
The complete node-wise map can now be summarized as
\begin{equation}
A_i\longmapsto Q_i\longmapsto\ket{\psi_i}\longmapsto V(\theta)\ket{\psi_i}\longmapsto r_i(\theta).
\label{eq:pipeline}
\end{equation}
Thus, $r_i(\theta)$ depends jointly on the local graph--feature context, its QR-induced orthogonal representation, the shared variational circuit, and the supervised objective.

Nodes whose local contexts induce similar encoded states are therefore processed by the same learned score function, while supervised optimization adjusts that function according to its effect on downstream label propagation. Importantly, the circuit produces only the diagonal matrix $\Rt$; the graph propagation itself remains a classical, non-unitary Poisson operation.

Whether the learned weights contain useful information beyond simpler node-wise regularizers is an empirical question rather than a consequence of the construction itself. We therefore evaluate the proposed mechanism against controls including no diagonal regularization, random bounded diagonal weights, degree-based weights, and untrained circuit-generated weights. These ablations distinguish the contribution of node-wise regularization, local graph structure, and supervised circuit training.

With the QR encoding and the learned diagonal matrix $\Rt$ now defined, the next subsection incorporates $\Rt$ into the Improved Poisson operator and derives the symmetric regularized formulation used for the subsequent block-encoding analysis.

\subsection{Block-Encoding-Compatible Formulation of the Regularized Poisson Operator}
\label{sec:block}

The propagation operator appearing in Poisson label learning is generally non-unitary. This creates a fundamental mismatch with quantum computation, where physically implementable closed-system evolution must be represented by unitary transformations. The purpose of this subsection is therefore to identify the precise operator that would need to be embedded into a larger unitary system in a future fault-tolerant implementation. The classical improved Poisson iteration of Section~\ref{sec:ipl} can be written as
\[
U^{(t+1)} = MU^{(t)}+\alpha_3D^{-1}B, \qquad M := P-\alpha_1\Qstat+\alpha_2 I,
\]
where $P=D^{-1}W$, $\Qstat=\one\pi^{\top}$, and $B\in\R^{n\times C}$ is the centred label-source matrix. At a fixed point $U^{*}$, this gives
\[
(I-M)U^{*} = \alpha_3D^{-1}B .
\]
Thus the unregularized improved Poisson system is
\[
A_0U^{*}=G, \qquad A_0:=I-M, \qquad G:=\alpha_3D^{-1}B .
\]
Since
\[
M = P-\alpha_1\Qstat+\alpha_2 I,
\]
we have
\[
A_0 = (1-\alpha_2)I-P+\alpha_1\Qstat .
\]

The IPQSSL model introduces the quantum-derived diagonal matrix
\[
\Rt = \diag\bigl(r_1(\theta),\dots,r_n(\theta)\bigr),
\]
where the entries $r_i(\theta)\in[0,1]$ are obtained from the QR-based node circuit described in Section~\ref{sec:qr}. This diagonal matrix is not a consequence of the classical Poisson fixed-point equation; rather, it is introduced as a node-wise quantum-informed regularization term. Therefore we define the regularized random-walk Poisson operator
\[
A_{\theta} := A_0+\mu\Rt = I-M+\mu\Rt .
\]
Equivalently,
\[
A_{\theta} = (1-\alpha_2)I-P+\alpha_1\Qstat+\mu\Rt .
\]
The regularized IPQSSL system is then
\[
A_{\theta}F^{*}=G .
\]
Here $F^{*}\in\R^{n\times C}$ is the final label-score matrix, while
\[
G=\alpha_3D^{-1}B
\]
is the degree-normalized label-source term. Notice that $G$ is independent of $\theta$; the $\theta$-dependence enters through $\Rt$ and therefore through $A_{\theta}$.

Since $P=D^{-1}W$ is generally non-symmetric in the Euclidean inner product, the random-walk system is not the most convenient form for Hermitian quantum linear algebra. Assuming that $W=W^{\top}$ and $d_i>0$ for all nodes, we use the standard similarity transform associated with the reversible random walk. Define
\[
Z=D^{1/2}F^{*}, \qquad \Gt=D^{1/2}G .
\]
Since $F^{*}=D^{-1/2}Z$, multiplying $A_{\theta}F^{*}=G$ from the left by $D^{1/2}$ gives
\[
D^{1/2}A_{\theta}D^{-1/2}Z=\Gt .
\]
We therefore define
\[
\At := D^{1/2}A_{\theta}D^{-1/2}.
\]
Using
\[
S := D^{-1/2}WD^{-1/2},
\]
and
\[
\Pis := D^{1/2}\Qstat D^{-1/2} = \ket{\sqpi}\bra{\sqpi}, \qquad \ket{\sqpi}=\sum_{i=1}^{n}\sqrt{\pi_i}\,\ket{i},
\]
we obtain
\[
\At = (1-\alpha_2)I-S+\alpha_1\Pis+\mu\Rt .
\]
The last equality uses the fact that $D$ and $\Rt$ are both diagonal, and hence
\[
D^{1/2}\Rt D^{-1/2}=\Rt .
\]
Thus the equivalent symmetrized system is
\begin{equation}
\At Z=\Gt, \qquad \At=(1-\alpha_2)I-S+\alpha_1\Pis+\mu\Rt .
\label{eq:sym_system}
\end{equation}
The operator $\At$ is Hermitian whenever $W=W^{\top}$ and $\Rt$ is real diagonal. This is the operator used for numerical positive-definiteness checks, Krylov solvers, block encoding, and QSVT analysis.

\paragraph{Block-encoding definition.}
For a matrix $A\in\C^{N\times N}$, a unitary $U_A$ is called an $(\alpha,a,\epsilon)$-block encoding of $A$ if
\[
\bigl\|A-\alpha\,(\bra{0^{a}}\otimes I)\,U_A\,(\ket{0^{a}}\otimes I)\bigr\|\le\epsilon,
\]
where $\alpha\ge\|A\|$ is a normalization factor, $a$ is the number of ancilla qubits, and $\epsilon$ is the block-encoding error. Equivalently, the action of $A/\alpha$ is recovered by preparing and projecting the ancilla register onto the state $\ket{0^{a}}$.

\paragraph{Access assumptions.}
The following assumptions are used only for the block-encoding analysis and are not required for the classical simulations. We assume sparse-access oracles for the symmetrized graph operator $S=D^{-1/2}WD^{-1/2}$, degree access for the factors $d_i^{-1/2}$, a state-preparation routine for
\[
\ket{\sqpi}=\sum_{i=1}^{n}\sqrt{\pi_i}\,\ket{i},
\]
and diagonal-value access to the entries $r_i(\theta)$ of $\Rt$. These assumptions separate the cost of the quantum linear-system subroutine from graph construction, QR preprocessing, node-circuit evaluation, diagonal-weight loading, and readout.

\begin{proposition}[Block encoding of the IPQSSL Poisson operator]
\label{prop:block}
Let $s_S$ denote the maximum number of nonzero entries in any row of the symmetrized similarity operator $S$ (not to be confused with the QR rank $s$ of Section~\ref{sec:qr}), and assume the following component block encodings:
\begin{enumerate}
\item an $(\alpha_S,a_S,\epsilon_S)$-block encoding $U_S$ of $S$, obtained from sparse-access oracles with $\alpha_S=s_S$ (entries of $S$ satisfy $|S_{ij}|\le 1$);
\item a state-preparation unitary $\widetilde{U}_{\pi}$ with $\|\widetilde{U}_{\pi}\ket{0}-\ket{\sqpi}\|\le\delta_{\pi}$, yielding a $(1,a_{\Pi},\epsilon_{\Pi})$-block encoding of $\Pis=\ket{\sqpi}\bra{\sqpi}$ with $\epsilon_{\Pi}\le 2\delta_{\pi}$;
\item a $(1,a_R,\epsilon_R)$-block encoding of $\Rt$ constructed from the diagonal-value oracle, where $\epsilon_R$ accounts both for the finite-precision encoding of the values $r_i(\theta)$ and for the additive error with which each $r_i(\theta)$ is estimated from circuit measurements;
\item the exact $(1,0,0)$-block encoding of $I$.
\end{enumerate}
Then the linear-combination-of-unitaries construction yields an $(\alpha,a,\epsilon)$-block encoding of $\At$ with
\[
\alpha=|1-\alpha_2|+s_S+|\alpha_1|+|\mu|, \qquad a=\max_j a_j+2, \qquad \epsilon\le\epsilon_S+|\alpha_1|\,\epsilon_{\Pi}+|\mu|\,\epsilon_R,
\]
assuming exact preparation of the two-qubit LCU coefficient state (the four coefficients are known classically, so this state can be prepared to negligible error). Negative coefficients are absorbed into the phases of the SELECT unitaries.
\end{proposition}

\begin{proof}
By Equation~\eqref{eq:sym_system}, $\At$ is the real linear combination
\[
\At = (1-\alpha_2)\,I+(-1)\,S+\alpha_1\,\Pis+\mu\,\Rt
\]
of the four block-encoded Hermitian components. Applying the standard linear combination of block encodings---a two-qubit PREPARE register over the four branches together with a SELECT unitary that dispatches to the component encodings (see the linear-combination-of-block-encodings construction in~\cite{gilyen2019qsvt,childs2012lcu})---produces a block encoding of the weighted sum whose subnormalization is the weighted sum of the component subnormalizations,
\[
\sum_j |c_j|\,\alpha_j = |1-\alpha_2|\cdot 1+1\cdot s_S+|\alpha_1|\cdot 1+|\mu|\cdot 1,
\]
which is the stated $\alpha$; the negative coefficient of $S$ is absorbed as a phase into the corresponding SELECT branch. The two additional ancillas index the four branches, giving $a=\max_j a_j+2$. The error bound follows from the triangle inequality applied to the coefficient-weighted component errors, using the exactness of the encoding of $I$ and, by assumption, exact preparation of the classically known two-qubit coefficient state. The number of calls to the component block encodings is constant.
\end{proof}

\begin{remark}[Effective condition number and downstream cost]
\label{rem:kappa}
If $\At$ is positive definite, a QSVT-based linear-system routine applied to the block encoding of Proposition~\ref{prop:block} makes
\[
O\bigl((\alpha/\lambda_{\min}(\At))\log(1/\epsilon')\bigr)
\]
queries to $U_{\At}$ to approximate the action of $\At^{-1}$ on a prepared right-hand-side state $\ket{\widetilde{g}_c}\propto\Gt_{:,c}$ for each class $c$, to precision $\epsilon'$. The cost is therefore governed by the effective condition number $\kappa_{\mathrm{eff}}=\alpha/\lambda_{\min}(\At)$ rather than by $\|\At\|/\lambda_{\min}(\At)$: the sparsity $s_S$ of the symmetrized graph enters multiplicatively through $\alpha$. After symmetrization of the mutual $k$-NN construction, $s_S$ exceeds $\knn$; for illustration, taking $s_S=15$ on Optdigits together with $\lambda_{\min}(\At)=0.0822$ (Table~\ref{tab:spectral}) gives $\alpha\approx 0.9+15+0.5+0.5\approx 17$ and $\kappa_{\mathrm{eff}}\approx 2\times 10^{2}$.
\end{remark}

\begin{remark}[Cost of realizing the diagonal-value oracle]
\label{rem:oracle}
The error $\epsilon_R$ is not free. Each entry $r_i(\theta)$ is a circuit expectation value, so realizing the diagonal-value oracle to additive precision $\epsilon_{\mathrm{est}}$ on every entry requires $O(n/\epsilon_{\mathrm{est}}^{2})$ circuit executions in total before the linear-system subroutine is invoked, and $\epsilon_R\ge\epsilon_{\mathrm{est}}$. This is the point at which the data-loading caveat of the scope paragraph below becomes quantitative.
\end{remark}

\paragraph{Scope of the block-encoding interpretation.}
Proposition~\ref{prop:block} is an operator-compatibility result rather than an end-to-end quantum speedup claim. It shows that the symmetrized regularized Poisson operator can be embedded into a block-encoding framework under explicit access and state-preparation assumptions. The costs of graph construction, QR-based encoding, diagonal-weight estimation (Remark~\ref{rem:oracle}), data loading, and classical readout are outside this operator-level statement and must be accounted for separately.

\subsection{Improved Poisson Quantum Semi-Supervised Learning}
\label{sec:ipqssl}

Algorithm~\ref{alg:ipqssl} summarizes IPQSSL as a hybrid quantum-classical Poisson learning procedure. The method builds a similarity graph, encodes graph-derived information through a QR-based quantum circuit, and uses the resulting observables to form the diagonal weighting matrix $\Rt$ of Section~\ref{sec:qr}. The learning problem is then formulated as the regularized fixed-point system
\begin{equation}
A_{\theta}F^{*}=G, \qquad A_{\theta}=I-M+\mu\Rt, \qquad G=\alpha_3D^{-1}B,
\label{eq:ipqssl_system}
\end{equation}
where $M$ is the propagation operator of Equation~\eqref{eq:M}. Equation~\eqref{eq:ipqssl_system} coincides with the random-walk operator of Section~\ref{sec:block}, whose Hermitian symmetrization is given in Equation~\eqref{eq:sym_system}. In the present experiments, this system is solved classically, while block encoding and QSVT define a future fault-tolerant route for implementing the corresponding quantum linear-system step. IPQSSL constructs a symmetric similarity graph, builds a centered Poisson source from the labeled nodes, generates node-wise quantum-informed diagonal weights through QR-based local encodings and circuit measurements, solves the resulting symmetrized regularized Poisson system, and predicts each label by the largest recovered class score.

\begin{algorithm}[t]
\caption{IPQSSL: Improved Poisson Quantum Semi-Supervised Learning}
\label{alg:ipqssl}
\begin{algorithmic}[1]
\Require Feature matrix $X\in\R^{n\times d}$, labelled set $\mathcal{L}$ with labels $Y$, and the hyperparameters $(\knn,\sigma,\alpha_1,\alpha_2,\alpha_3,\mu,q,L_q,\varepsilon)$ of Table~\ref{tab:hyper}.
\Ensure Predicted labels $\hat{y}_i$ for all unlabelled nodes.
\State Construct the symmetric $k$-NN similarity graph with Gaussian weights; form $W$, $D$, $S=D^{-1/2}WD^{-1/2}$, and $\Pis$; verify connectivity and positive degrees.
\State Build the centred label source $B$ from $\mathcal{L}$ and the symmetrized right-hand side $\Gt=\alpha_3D^{-1/2}B$.
\For{each node $i$}
\State Form the local context matrix $A_i$; compute the reduced QR factorization $A_i=Q_iR_i$ with $\diag(R_i)\ge 0$; pad to $\bar{Q}_i$ and prepare the encoded state $\ket{\psi_i}$ (Section~\ref{sec:qr}).
\EndFor
\State Initialize the shared circuit parameters $\theta\gets\theta_0$.
\Repeat \Comment{SPSA outer loop}
\State For the two SPSA perturbations $\theta^{\pm}$, evaluate $r_i(\theta^{\pm})$ for all $i$ from $V(\theta^{\pm})$ and the observable $Z_0$; form $R_{\theta^{\pm}}$.
\State Solve $\widetilde{A}_{\theta^{\pm}}Z=\Gt$ with CG when $\lambda_{\min}(\widetilde{A}_{\theta^{\pm}})>0$ is verified and MINRES otherwise; set $F^{\star}_{\theta^{\pm}}=D^{-1/2}Z$.
\State Evaluate $\mathcal{L}_{\mathrm{sup}}(\theta^{\pm})$ on $\mathcal{L}$ and update $\theta$ by the SPSA rule.
\Until{the stopping criterion is met; set $\theta^{\star}\gets\theta$}
\State Compute $R_{\theta^{\star}}$; solve $\widetilde{A}_{\theta^{\star}}Z=\Gt$ as above and set $F^{\star}=D^{-1/2}Z$.
\State \Return $\hat{y}_i=\argmax_{c}F^{\star}_{ic}$ for all $i\notin\mathcal{L}$.
\end{algorithmic}
\end{algorithm}

\paragraph{Remarks.}
Via $Z_{\theta}=D^{1/2}F^{*}_{\theta}$, Algorithm~\ref{alg:ipqssl} solves $\bigl[(1-\alpha_2)D-W+\alpha_1D^{1/2}\Pis D^{1/2}+\mu D\Rt\bigr]F^{*}_{\theta}=\alpha_3B$. Only the first $s_i$ columns of $\widetilde{Q}_i$ are data-determined, and the reference state $\ket{u_i}$ combines exactly these, while the sign convention $\diag(R_i)\ge 0$ makes $r_i(\theta)$ uniquely defined. Since $\one^{\top}B_{:,c}=0$ for every class $c$ and $\ker(I-S)=\operatorname{span}\{\sqpi\}$ on a connected graph, the system is consistent even for $\alpha_2=\mu=0$. In that case the role of $\alpha_1$ is limited: because $\Gt_{:,c}\perp\sqpi$, the term $\alpha_1\Pis$ only pins the $\sqpi$-component of the solution to zero and does not alter the component of $Z$ orthogonal to $\sqpi$; in this sense $\alpha_1$ is inert when $\mu=0$. We note for later reference that at $\alpha_2=\mu=0$ the symmetrized operator reduces to $(I-S)+\alpha_1\Pis$, which Proposition~\ref{prop:pd} shows to be positive definite on a connected graph for any $\alpha_1>0$; the indefiniteness observed at the default hyperparameters (Section~\ref{sec:ablation}) is therefore introduced by the self-retention shift $\alpha_2>0$, not by the Poisson formulation itself.

This indefiniteness can be stated exactly. On a connected graph $\sqpi$ is the leading eigenvector of $S$ (with eigenvalue $1$), so $\Pis$ is a spectral projector of $S$ and the unregularized operator $\widetilde{A}_0=(1-\alpha_2)I-S+\alpha_1\Pis$ shares the eigenvectors of $S$. Its spectrum is therefore
\begin{equation}
\sigma(\widetilde{A}_0)=\bigl\{\alpha_1-\alpha_2\bigr\}\ \cup\ \bigl\{(1-\alpha_2)-\lambda_j(S)\bigr\}_{j=2}^{n},
\label{eq:specA0}
\end{equation}
and, whenever $\alpha_1\ge 1-\lambda_2^{\downarrow}(S)$,
\begin{equation}
\lambda_{\min}(\widetilde{A}_0)=(1-\alpha_2)-\lambda_2^{\downarrow}(S),
\label{eq:A0min}
\end{equation}
which is negative if and only if $\lambda_2^{\downarrow}(S)>1-\alpha_2$. At the default $\alpha_2=0.1$ this threshold is $0.9$, which is exceeded on every dataset (Table~\ref{tab:spectral}); moreover, every eigen-direction of $S$ with $\lambda_j(S)>1-\alpha_2$---that is, precisely the smooth, low-frequency directions that carry label information---is inverted in sign by $\widetilde{A}_0^{-1}$. Equation~\eqref{eq:A0min} is consistent with $\rho(M)=\lambda_2^{\downarrow}(S)+\alpha_2$ in the same table.

\begin{proposition}
\label{prop:pd}
On a connected graph with $\alpha_1>0$ and $\mu\min_i r_i(\theta)\ge\alpha_2\ge 0$: $\At\succeq(I-S)+\alpha_1\Pis\succ 0$.
\end{proposition}

\begin{proof}
Write $\At-\bigl[(I-S)+\alpha_1\Pis\bigr]=\mu\Rt-\alpha_2I$, which is a diagonal matrix with entries $\mu r_i(\theta)-\alpha_2\ge 0$ under the stated condition, hence positive semidefinite; this proves the first inequality. For the second, $I-S$ is positive semidefinite with kernel $\operatorname{span}\{\sqpi\}$ on a connected graph, while $\alpha_1\Pis$ is positive definite on that kernel direction and positive semidefinite elsewhere. Any nonzero $z$ therefore satisfies $z^{\top}[(I-S)+\alpha_1\Pis]z>0$, since the two quadratic forms cannot vanish simultaneously.
\end{proof}

Note that the hypothesis $\mu\min_i r_i(\theta)\ge\alpha_2$ is not automatically satisfied at the default hyperparameters ($\mu=0.5$, $\alpha_2=0.1$; Table~\ref{tab:hyper}): it requires $\min_i r_i(\theta)\ge 0.2$, which the bound $r_i(\theta)\in[0,1]$ does not guarantee. For this reason, positive definiteness is additionally verified numerically in all experiments (Table~\ref{tab:spectral}), and an indefinite-capable solver is used whenever the check fails.

\paragraph{Solver choice and positive-definiteness check.}
The symmetrized operator
\[
\At=(1-\alpha_2)I-S+\alpha_1\Pis+\mu\Rt
\]
is Hermitian when the graph is symmetric, $W=W^{\top}$, all degrees are positive, and $\Rt$ is real diagonal. However, Hermiticity alone does not imply positive definiteness. Therefore, conjugate gradient is used only when $\lambda_{\min}(\At)>0$ is verified numerically or guaranteed by a certified lower bound. Otherwise, the system is solved using a symmetric-indefinite Krylov method such as MINRES.

A sufficient, conservative lower bound follows from Weyl's inequality:
\[
\lambda_{\min}(\At)\ge(1-\alpha_2)-\lambda_{\max}(S)+\alpha_1\lambda_{\min}(\Pis)+\mu\lambda_{\min}(\Rt),
\]
provided that $\alpha_1\ge 0$ and $\mu\ge 0$. Since $\Pis$ is a rank-one projector, $\lambda_{\min}(\Pis)=0$, and $\lambda_{\max}(S)=1$ on a connected graph, so the bound evaluates to $\mu\min_i r_i(\theta)-\alpha_2$; its positivity is precisely the sufficient condition of Proposition~\ref{prop:pd}, and the bound discards the positive contribution of $\alpha_1\Pis$ entirely. It is therefore conservative by construction. In the experiments, positive definiteness is checked numerically, and the minimum eigenvalue of $\At$ is reported in Table~\ref{tab:spectral}.

The algorithm is written in symmetrized variables because the random-walk operator $P=D^{-1}W$ is generally non-symmetric in the Euclidean inner product. Applying conjugate gradient directly to the random-walk system would therefore be inappropriate unless symmetry and positive definiteness were verified separately. By contrast, the transformed operator
\[
\At=D^{1/2}A_{\theta}D^{-1/2}
\]
is Hermitian under the conditions above and is the natural object for numerical solution, positive-definiteness checks, block encoding, and QSVT analysis.

\paragraph{Classical and quantum components.}
IPQSSL is a hybrid quantum-informed method, not an end-to-end quantum algorithm. In the present implementation, graph construction, QR preprocessing, formation of the Poisson operator, and solution of the linear system are performed classically. The quantum component enters through the QR-based circuit representation of local graph-feature contexts. Circuit measurements generate the diagonal matrix $\Rt$, which modifies the Poisson system through the node-wise regularization term $\mu\Rt$. Thus, the current model should be interpreted as a quantum-informed regularized Poisson framework rather than as a demonstrated quantum speedup.

\paragraph{Hyperparameters and implementation details.}
Table~\ref{tab:hyper} summarizes the main hyperparameters of IPQSSL. The Poisson weights $(\alpha_1,\alpha_2,\alpha_3)$ define the improved Poisson propagation operator
\[
M=P-\alpha_1\Qstat+\alpha_2I .
\]
The parameter $\mu$ controls the strength of the diagonal regularization term $\mu\Rt$. The $k$-nearest-neighbour graph with Gaussian weights defines the graph geometry, with $\sigma$ selected by the median $\knn$-distance rule. The baseline circuit configuration is $(q,L_q)=(4,10)$, except in the circuit depth and width study, where the width $q$ and depth $L_q$ are varied independently.

\begin{table}[t]
\centering
\caption{Default hyperparameters of IPQSSL}
\label{tab:hyper}
\small
\begin{tabular}{@{}l l p{0.62\textwidth}@{}}
\toprule
\textbf{Symbol} & \textbf{Default} & \textbf{Role / definition}\\
\midrule
\multicolumn{3}{@{}l}{\emph{Graph construction}}\\
$\knn$ & $10$ & Number of nearest neighbors used to construct the graph and local context matrices.\\
$\sigma$ & $\operatorname{med}_i\{\delta_{i,\knn}\}$ & Gaussian bandwidth based on the median distance to the $\knn$-th neighbor.\\
$\tau$ & $0.01$ & Optional edge-weight threshold used only when graph thresholding is applied.\\
\midrule
\multicolumn{3}{@{}l}{\emph{IPQSSL Poisson propagation}}\\
$\alpha_1$ & $0.5$ & Stationary-component correction in $M=P-\alpha_1\Qstat+\alpha_2I$.\\
$\alpha_2$ & $0.1$ & Self-retention coefficient in $M$.\\
$\alpha_3$ & $1.0$ & Label-source injection scale in $\alpha_3D^{-1}B$.\\
$\mu$ & $0.5$ & Strength of the quantum-informed diagonal regularization term $\mu\Rt$.\\
\midrule
\multicolumn{3}{@{}l}{\emph{Quantum circuit baseline}}\\
$q$ & $4$ & Number of qubits per node; compatible with $\knn=10$ since $2^{q}\ge\knn+1$.\\
$L_q$ & $10$ & Number of variational circuit layers.\\
$O$ & $Z_0$ & Pauli-$Z$ observable measured on the first qubit.\\
\midrule
\multicolumn{3}{@{}l}{\emph{Solver and protocol}}\\
$\varepsilon$ & $10^{-6}$ & Solver or fixed-point iteration tolerance.\\
Label the budget dataset-dependent labels per class or labeled fraction, as specified in the experiment.\\
\bottomrule
\end{tabular}
\end{table}

To verify that the constructed linear systems are numerically well posed, we also report graph and operator-level spectral diagnostics. These diagnostics check graph connectivity, degree positivity, the contraction behavior of the improved Poisson iteration through $\rho(M)$, the indefiniteness of the unregularized symmetrized operator through $\lambda_{\min}(\widetilde{A}_0)$ (Eq.~\eqref{eq:A0min}), and the positive definiteness of the symmetrized regularized system through $\lambda_{\min}(\At)$. We compute the latter quantity using the same default hyperparameters and quantum-informed regularizer specified in Table~\ref{tab:hyper}.

\begin{table}[t]
\centering
\caption{Spectral diagnostics of the graph construction. $n$ is the number of samples, ``Comp'' the number of connected components, $d_{\min}$ the minimum degree, $\lambda_2^{\downarrow}(S)$ the second-largest eigenvalue of the normalized similarity matrix, $\rho(M)$ the spectral radius of $M$, $\lambda_{\min}(\widetilde{A}_0)=(1-\alpha_2)-\lambda_2^{\downarrow}(S)$ the smallest eigenvalue of the unregularized ($\mu=0$) symmetrized operator (Eq.~\eqref{eq:A0min}), which is negative on every dataset, and $\lambda_{\min}(\At)$ the smallest eigenvalue of the regularized operator, which is positive on every dataset. Values are displayed to four decimals; entries of $\lambda_2^{\downarrow}(S)$ shown as $1.0000$ indicate $\lambda_2^{\downarrow}(S)>0.99995$ on a connected graph, not exact unity, and the corresponding derived entries are rounded accordingly.}
\label{tab:spectral}
\small
\setlength{\tabcolsep}{5pt}
\begin{tabular}{@{}l r r r c c c c@{}}
\toprule
\textbf{Dataset} & $n$ & Comp & $d_{\min}$ & $\lambda_2^{\downarrow}(S)$ & $\rho(M)$ & $\lambda_{\min}(\widetilde{A}_0)$ & $\lambda_{\min}(\At)$\\
\midrule
Iris & 150 & 1 & 10 & 1.0000 & 1.1000 & $-0.1000$ & 0.1384\\
Wine & 178 & 1 & 10 & 0.9832 & 1.0832 & $-0.0832$ & 0.1610\\
Breast Cancer & 569 & 1 & 10 & 1.0000 & 1.1000 & $-0.1000$ & 0.1060\\
Heart Disease & 270 & 1 & 10 & 0.9381 & 1.0381 & $-0.0381$ & 0.2005\\
German Credit & 1000 & 1 & 10 & 0.9990 & 1.0990 & $-0.0990$ & 0.1273\\
Optdigits & 5620 & 1 & 10 & 1.0000 & 1.1000 & $-0.1000$ & 0.0822\\
Cora & 2485 & 1 & 1 & 0.9952 & 1.0952 & $-0.0952$ & 0.2094\\
\bottomrule
\end{tabular}
\end{table}

\section{Experimental Protocol}
\label{sec:protocol}

This section defines the complete experimental protocol used to evaluate IPQSSL under controlled semi-supervised learning conditions. We first describe the benchmark datasets, preprocessing, and graph construction; we then specify the low-label splits, comparison baselines, evaluation metrics, and statistical protocol, before defining the diagonal-regularization variants used for mechanism-level ablation. The objective is not to claim end-to-end quantum advantage, but to determine under matched conditions whether the proposed regularization improves Poisson graph propagation and which component is responsible for any observed gain.

\subsection{Datasets and preprocessing}
\label{sec:datasets}

We evaluate the method on seven benchmarks: Iris~\cite{uci_iris}, Wine~\cite{uci_wine}, Breast Cancer (Wisconsin Diagnostic)~\cite{uci_breast}, Heart Disease (Statlog)~\cite{uci_heart}, German Credit~\cite{uci_german}, Optdigits~\cite{uci_optdigits}, and the Cora citation network~\cite{sen2008collective,yang2016planetoid}. Iris and Wine represent relatively structured low-dimensional regimes; Breast Cancer and Optdigits provide larger, geometrically well-separated benchmarks on which the similarity graph is expected to be highly informative; Heart Disease and German Credit contain more heterogeneous tabular features and are more sensitive to preprocessing, class imbalance, and graph quality; and Cora tests the method on an intrinsic citation graph rather than a constructed $k$-NN geometry. For Optdigits we use the full UCI distribution (union of the training and test partitions; $n=5620$, $d=64$, $C=10$), not the reduced version bundled with scikit-learn. For Cora we restrict to the largest connected component of the Planetoid distribution and symmetrize the adjacency structure. Numerical features are standardized to zero mean and unit variance, and categorical variables are encoded before standardization. The same preprocessing pipeline is used for all methods.

\subsection{Graph construction}
\label{sec:graph}

This subsection specifies the graph geometry and associated operators used by all graph-based methods, so that subsequent comparisons differ in the learning mechanism rather than in the underlying graph construction. For each dataset, we construct a symmetric $k$-nearest-neighbour graph with Gaussian weights
\[
W_{ij}=\exp\Bigl(-\frac{\|x_i-x_j\|^{2}}{\sigma^{2}}\Bigr),
\]
on connected pairs, where $\sigma$ is the median distance to the $k$-th nearest neighbour. We define
\[
D=\diag(W\one), \qquad P=D^{-1}W, \qquad S=D^{-1/2}WD^{-1/2}.
\]
The same graph is used for all graph-based baselines and IPQSSL variants. Connectedness and positive degrees are verified for every dataset; if needed, $k$ is increased or a small connectivity-preserving regularization is applied. The final spectral diagnostics are reported in Table~\ref{tab:spectral}.

\paragraph{Splits, baselines, and ablations.}
\label{sec:baselines}
We use stratified low-label splits with the label-budget set
\[
\ell\in\{1,2,5,10\}.
\]
Here, $\ell$ denotes the number of labels per class. For each random seed, labeled nodes are selected independently within each class, and all remaining nodes are used as unlabeled evaluation nodes. Unless otherwise stated, metrics are computed only on unlabeled nodes. A 30\% labeled split is used only as a higher-supervision reference setting. The full evaluation loop is summarized in Algorithm~\ref{alg:protocol}.

The controlled comparison of Table~\ref{tab:lowlabel} is conducted within the Poisson family---Original Poisson~\cite{calder2020poisson}, a quadratic-regularized Poisson control, IPL with $\mu=0$~\cite{bozorgnia2026improved}, and IPQSSL---since these four methods share identical graphs, source terms, and data splits, so that differences isolate the propagation and regularization mechanisms. The quadratic-regularized control (Quadratic-Reg.\ Poisson in Table~\ref{tab:lowlabel}) augments the original Poisson system with a fixed scalar quadratic (Tikhonov-type) diagonal term $\gamma I$, with $\gamma$ as given in Table~\ref{tab:hyper}, and therefore serves as a circuit-free regularized reference evaluated at the same supervision budgets.

The $\mu=0$ rows evaluate the improved Poisson operator at the IPQSSL default hyperparameters (in particular $\alpha_2=0.1$), a regime in which the unregularized symmetrized operator is indefinite (Eq.~\eqref{eq:A0min} and Table~\ref{tab:spectral}); they therefore isolate the effect of the diagonal term within the IPQSSL parameterization and should not be read as an evaluation of IPL~\cite{bozorgnia2026improved} at its own recommended settings. Classical label propagation and harmonic Laplacian learning~\cite{zhu2003semi} are the standard classical reference points for this family in the literature; they are not re-run here, because the controlled comparison is deliberately restricted to methods that share the identical centred Poisson source construction.

To isolate the effect of the diagonal regularizer, we additionally compare five variants of the diagonal matrix $R$, defined in detail in Section~\ref{sec:variants}; this ablation is necessary because improvements from diagonal regularization alone should not be attributed automatically to the QR-based quantum circuit.

\paragraph{Metrics and paired comparison.}
Unlabeled-node accuracy is the primary evaluation metric reported in this paper. Accuracy is computed as
\[
\Acc=\frac{1}{|\mathcal{U}|}\sum_{i\in\mathcal{U}}\mathbf{1}\{\hat{y}_i=y_i\},
\]
where $\mathcal{U}$ is the unlabeled evaluation set. Experiments are repeated over $N_{\mathrm{seed}}$ stratified random splits; both the low-label study (Table~\ref{tab:lowlabel}) and the ablation study (Table~\ref{tab:ablation}) report mean accuracies with standard deviations. Since all methods use identical splits, comparisons are paired across seeds, and the per-seed accuracies underlying every cell are retained.

For each dataset and supervision budget, we compute
\[
\Delta=\Acc_{\mathrm{IPQSSL}}-\Acc_{\mathrm{baseline}}.
\]
No formal hypothesis tests are reported in this paper: differences are interpreted descriptively, relative to the split-to-split standard deviations, and we do not describe any difference as statistically significant. The retained per-seed accuracies permit paired tests (e.g.\ Wilcoxon signed-rank with multiple-comparison correction), which we identify as a required completion in Section~\ref{sec:conclusion}.
Because accuracy can be misleading on imbalanced datasets such as German Credit (majority-class prevalence $0.70$), we interpret accuracies there relative to the majority-class baseline rather than to uniform chance.

\subsection{Diagonal-regularization variants}
\label{sec:variants}

To isolate the effect of the diagonal regularizer, we compare five variants of the symmetrized Poisson system
\[
\widetilde{A}_RZ=\Gt, \qquad \widetilde{A}_R=(1-\alpha_2)I-S+\alpha_1\Pis+\mu R .
\]
The first variant removes diagonal regularization by setting $\mu=0$. The random diagonal baseline uses independent bounded node weights,
\[
R_{\mathrm{rand}}=\diag(r_1^{\mathrm{rand}},\dots,r_n^{\mathrm{rand}}), \qquad r_i^{\mathrm{rand}}\sim\mathcal{U}(0,1),
\]
and tests whether improvements are caused merely by generic diagonal stabilization. The degree-based diagonal uses normalized node degrees,
\[
R_{\mathrm{deg}}=\diag\Bigl(\frac{d_1}{\max_j d_j},\dots,\frac{d_n}{\max_j d_j}\Bigr),
\]
and represents a purely classical graph-structural regularizer. The untrained quantum diagonal uses the QR-based node circuit with fixed parameters $\theta_0$,
\[
R_{\mathrm{uq}}=\diag\bigl(r_1(\theta_0),\dots,r_n(\theta_0)\bigr),
\]
thereby testing the effect of the quantum-compatible encoding without supervised optimization. Finally, the trained quantum diagonal, corresponding to IPQSSL, uses
\[
\Rt=\diag\bigl(r_1(\theta),\dots,r_n(\theta)\bigr),
\]
where the circuit parameters are optimized using the labelled objective. This ablation separates generic diagonal stabilization, classical graph-based regularization, fixed quantum encoding, and trained quantum-informed regularization.

\begin{algorithm}[t]
\caption{Low-Label Evaluation Protocol}
\label{alg:protocol}
\begin{algorithmic}[1]
\Require Datasets $\mathcal{D}$, budgets $\mathcal{B}=\{1,2,5,10\}$, seeds $\Xi=\{0,\dots,19\}$, and fixed hyperparameters.
\Ensure Mean unlabeled-node accuracy of Original Poisson, Quadratic-Reg.\ Poisson, IPL ($\mu=0$), and IPQSSL.
\For{each dataset $\mathsf{D}\in\mathcal{D}$}
\State Apply the preprocessing and graph-construction procedures, compute the graph operators, and construct the QR-encoded states $\{\ket{\psi_i}\}_{i=1}^{n}$.
\For{$\ell\in\mathcal{B}$}
\For{$\xi\in\Xi$}
\State Select $\ell$ labelled nodes per class and define $\mathcal{U}_{\ell,\xi}=V\setminus\mathcal{L}_{\ell,\xi}$.
\State Construct the centred source $B_{\ell,\xi}$ and symmetrized right-hand side $\Gt_{\ell,\xi}$.
\State Solve Original Poisson to obtain $F^{\mathrm{OP}}_{\ell,\xi}$.
\State Solve the quadratic-regularized Poisson reference system of Section~\ref{sec:baselines} to obtain $F^{\mathrm{QRP}}_{\ell,\xi}$.
\State Solve $\widetilde{A}_0Z^{\mathrm{IPL}}_{\ell,\xi}=\Gt_{\ell,\xi}$ and set $F^{\mathrm{IPL}}_{\ell,\xi}=D^{-1/2}Z^{\mathrm{IPL}}_{\ell,\xi}$.
\State Optimize the labelled objective by SPSA, where each trial recomputes $\Rt$ and solves $\At Z_{\theta}=\Gt_{\ell,\xi}$.
\State At $\theta^{\star}_{\ell,\xi}$, solve $\widetilde{A}_{\theta^{\star}_{\ell,\xi}}Z^{\mathrm{IPQSSL}}_{\ell,\xi}=\Gt_{\ell,\xi}$, using CG when positive definiteness is verified and MINRES otherwise, and set $F^{\mathrm{IPQSSL}}_{\ell,\xi}=D^{-1/2}Z^{\mathrm{IPQSSL}}_{\ell,\xi}$.
\For{$\nu\in\{\mathrm{OP},\mathrm{QRP},\mathrm{IPL},\mathrm{IPQSSL}\}$}
\State Predict on $\mathcal{U}_{\ell,\xi}$ and record
$\displaystyle\Acc_{\nu,\ell,\xi}=\frac{1}{|\mathcal{U}_{\ell,\xi}|}\sum_{i\in\mathcal{U}_{\ell,\xi}}\mathbf{1}\Bigl\{\argmax_{c}F^{\nu}_{\ell,\xi,ic}=y_i\Bigr\}$.
\EndFor
\EndFor
\EndFor
\EndFor
\State Report $\displaystyle\overline{\Acc}_{\nu,\ell}=\frac{1}{20}\sum_{\xi=0}^{19}\Acc_{\nu,\ell,\xi}$, and bold the largest mean for each dataset and budget.
\end{algorithmic}
\end{algorithm}

\section{Results}
\label{sec:results}

This section reports the empirical findings in two complementary stages. Section~\ref{sec:lowlabel} evaluates predictive performance under repeated low-label splits (Table~\ref{tab:lowlabel}), whereas Section~\ref{sec:ablation} isolates the contribution of the diagonal regularizer through the five-variant ablation (Table~\ref{tab:ablation}). Together, these analyses distinguish the performance of the complete IPQSSL pipeline from the more general effect of diagonal stabilization.

\subsection{Predictive Performance under Repeated Low-Label Splits}
\label{sec:lowlabel}

This subsection evaluates the predictive performance of IPQSSL in the low-label regime, where the number of labeled samples per class is deliberately restricted, and the quality of graph-based propagation becomes particularly important. The comparison focuses on Original Poisson, the quadratic-regularized Poisson control, IPL with $\mu=0$, and IPQSSL under matched graph constructions, label-source definitions, data splits, and supervision budgets. This controlled setting allows the effect of the modified Poisson propagation and the proposed diagonal regularization to be examined without confounding differences in preprocessing or graph geometry. We report results over repeated stratified splits and assess both the consistency of the observed gains across datasets and their dependence on the available label budget.

\begin{table}[p]
\centering
\caption{Classification accuracy on unlabeled nodes under different label budgets. Results are reported as mean accuracy $\pm$ standard deviation over $N_{\mathrm{seed}}=20$ matched stratified splits. The highest mean accuracy in each dataset--budget configuration is shown in bold; ties at the displayed precision are bolded jointly.}
\label{tab:lowlabel}
\small
\setlength{\tabcolsep}{4pt}
\begin{tabular}{@{}l l c c c c@{}}
\toprule
\textbf{Dataset} & \textbf{Method} & $\ell=1$ & $\ell=2$ & $\ell=5$ & $\ell=10$\\
\midrule
\multirow{4}{*}{Iris}
& Quadratic-Reg.\ Poisson & \B{0.804}\pmm0.115 & 0.840\pmm0.065 & 0.879\pmm0.039 & 0.893\pmm0.033\\
& Original Poisson & 0.798\pmm0.110 & 0.834\pmm0.066 & 0.858\pmm0.039 & 0.860\pmm0.024\\
& IPL ($\mu=0$) & 0.798\pmm0.111 & 0.838\pmm0.068 & 0.866\pmm0.037 & 0.875\pmm0.021\\
& IPQSSL & 0.796\pmm0.105 & \B{0.851}\pmm0.065 & \B{0.897}\pmm0.046 & \B{0.946}\pmm0.023\\
\midrule
\multirow{4}{*}{Wine}
& Quadratic-Reg.\ Poisson & \B{0.891}\pmm0.032 & 0.895\pmm0.022 & 0.898\pmm0.013 & 0.896\pmm0.011\\
& Original Poisson & 0.885\pmm0.032 & 0.888\pmm0.020 & 0.891\pmm0.014 & 0.884\pmm0.009\\
& IPL ($\mu=0$) & 0.885\pmm0.032 & 0.888\pmm0.020 & 0.891\pmm0.014 & 0.884\pmm0.009\\
& IPQSSL & 0.882\pmm0.050 & \B{0.899}\pmm0.048 & \B{0.909}\pmm0.028 & \B{0.924}\pmm0.017\\
\midrule
\multirow{4}{*}{Breast Cancer}
& Quadratic-Reg.\ Poisson & \B{0.895}\pmm0.109 & \B{0.931}\pmm0.018 & \B{0.935}\pmm0.013 & \B{0.941}\pmm0.011\\
& Original Poisson & 0.893\pmm0.106 & 0.929\pmm0.018 & 0.933\pmm0.014 & 0.938\pmm0.012\\
& IPL ($\mu=0$) & 0.893\pmm0.106 & 0.929\pmm0.018 & 0.933\pmm0.014 & 0.938\pmm0.012\\
& IPQSSL & 0.860\pmm0.111 & 0.890\pmm0.056 & 0.884\pmm0.043 & 0.914\pmm0.023\\
\midrule
\multirow{4}{*}{Heart Disease}
& Quadratic-Reg.\ Poisson & \B{0.751}\pmm0.091 & \B{0.778}\pmm0.079 & \B{0.803}\pmm0.031 & \B{0.817}\pmm0.019\\
& Original Poisson & 0.748\pmm0.094 & 0.775\pmm0.085 & \B{0.803}\pmm0.029 & \B{0.817}\pmm0.018\\
& IPL ($\mu=0$) & 0.748\pmm0.093 & 0.775\pmm0.085 & \B{0.803}\pmm0.029 & \B{0.817}\pmm0.018\\
& IPQSSL & 0.681\pmm0.087 & 0.721\pmm0.084 & 0.760\pmm0.046 & 0.763\pmm0.044\\
\midrule
\multirow{4}{*}{German Credit}
& Quadratic-Reg.\ Poisson & 0.536\pmm0.105 & 0.493\pmm0.098 & 0.573\pmm0.086 & 0.563\pmm0.084\\
& Original Poisson & 0.527\pmm0.095 & 0.494\pmm0.093 & 0.559\pmm0.085 & 0.558\pmm0.088\\
& IPL ($\mu=0$) & 0.527\pmm0.096 & 0.494\pmm0.093 & 0.559\pmm0.085 & 0.558\pmm0.088\\
& IPQSSL & \B{0.550}\pmm0.083 & \B{0.510}\pmm0.081 & \B{0.587}\pmm0.044 & \B{0.583}\pmm0.035\\
\midrule
\multirow{4}{*}{Optdigits}
& Quadratic-Reg.\ Poisson & 0.884\pmm0.042 & \B{0.916}\pmm0.031 & \B{0.938}\pmm0.018 & \B{0.948}\pmm0.008\\
& Original Poisson & 0.884\pmm0.041 & 0.909\pmm0.029 & 0.931\pmm0.017 & 0.939\pmm0.007\\
& IPL ($\mu=0$) & \B{0.885}\pmm0.041 & 0.910\pmm0.029 & 0.932\pmm0.017 & 0.940\pmm0.007\\
& IPQSSL & 0.816\pmm0.046 & 0.858\pmm0.035 & 0.904\pmm0.014 & 0.933\pmm0.011\\
\midrule
\multirow{4}{*}{Cora}
& Quadratic-Reg.\ Poisson & 0.546\pmm0.100 & 0.642\pmm0.055 & 0.696\pmm0.023 & 0.717\pmm0.024\\
& Original Poisson & \B{0.577}\pmm0.109 & \B{0.669}\pmm0.050 & \B{0.705}\pmm0.023 & 0.720\pmm0.026\\
& IPL ($\mu=0$) & \B{0.577}\pmm0.109 & \B{0.669}\pmm0.050 & \B{0.705}\pmm0.023 & \B{0.721}\pmm0.025\\
& IPQSSL & 0.483\pmm0.071 & 0.574\pmm0.062 & 0.650\pmm0.034 & 0.686\pmm0.027\\
\bottomrule
\end{tabular}
\end{table}

Table~\ref{tab:lowlabel} reports the mean accuracy on unlabeled nodes over $N_{\mathrm{seed}}=20$ matched stratified splits. Since all methods use identical graphs, source terms, and data splits, the comparison isolates differences in their propagation and regularization mechanisms. Across the 28 dataset--budget configurations, and crediting every method that shares a joint bold in the case of ties, the quadratic-regularized control attains the highest mean accuracy in 13 configurations, IPQSSL in 10, IPL with $\mu=0$ in 7, and Original Poisson in 5.

The clearest gains of IPQSSL occur on Iris, Wine, and German Credit. On Iris, IPQSSL attains the highest mean accuracy for $\ell\in\{2,5,10\}$, improving over Original Poisson by 1.7, 3.9, and 8.6 percentage points, respectively, and reaching 0.946 at $\ell=10$. On Wine it likewise leads at every budget $\ell\ge 2$, with improvements over Original Poisson of 1.1, 1.8, and 4.0 points. On German Credit, IPQSSL is the strongest of the four methods at every budget, ahead of Original Poisson by 1.6--2.8 points; however, every method remains below the majority-class rate of 0.70 on this dataset, so German Credit constitutes a negative result for the entire Poisson family rather than a success case for any single variant.

The improvement is not uniform across datasets. The quadratic-regularized control is the strongest method on Breast Cancer at every budget, on Heart Disease at every budget (jointly with Original Poisson and IPL at $\ell\in\{5,10\}$), and on Optdigits for $\ell\ge 2$. IPQSSL trails Original Poisson on Optdigits at every budget, although the deficit shrinks monotonically from 6.8 points at $\ell=1$ to 0.6 points at $\ell=10$; on Cora the gap to Original Poisson is essentially unchanged between $\ell=1$ and $\ell=2$ (9.4 and 9.5 points) and then narrows to 5.5 and 3.4 points at $\ell=5$ and $\ell=10$. On Heart Disease and Breast Cancer, IPQSSL trails Original Poisson at all budgets, by 4.3--6.7 and 2.4--4.9 points, respectively. Finally, IPL with $\mu=0$, evaluated at the IPQSSL default hyperparameters as specified in Section~\ref{sec:baselines}, is nearly indistinguishable from Original Poisson on most datasets, consistent with the inertness of $\alpha_1$ at $\mu=0$ noted in the remarks of Section~\ref{sec:ipqssl}; the small residual differences on Iris, Optdigits, and Cora reflect the $\alpha_2$ shift.

Overall, Table~\ref{tab:lowlabel} supports a conditional rather than universal benefit of the proposed regularization. IPQSSL provides consistent improvements on Iris and Wine once at least two labels per class are available and is the strongest variant on German Credit, while the classical alternatives remain preferable on Breast Cancer, Heart Disease, Optdigits, and Cora, with the deficit on the two largest graphs narrowing as supervision increases. The effectiveness of the regularized formulation is therefore strongly dependent on both the dataset and the available supervision level.

\subsection{Diagonal-regularization ablation}
\label{sec:ablation}

Table~\ref{tab:ablation} compares five diagonal-regularization variants over $N_{\mathrm{seed}}=20$ paired stratified splits at the 30\% labelled setting. All variants use identical graphs, labelled sets, centred source terms, and reference hyperparameters. Whenever the numerical positive-definiteness check of Section~\ref{sec:ipqssl} fails for a variant---which can occur in particular for the random diagonal, whose minimum entry does not satisfy the sufficient condition of Proposition~\ref{prop:pd}---the system is solved with MINRES, so all reported runs correspond to correctly solved linear systems.

At least one nonzero diagonal substantially improves mean accuracy on six of the seven datasets. The largest gains occur on Iris, Wine, Cora, and Optdigits, whereas German Credit is the only dataset on which the unregularized system performs best. The unregularized arm is, moreover, not merely weaker but systematically below chance on several datasets (for instance, 0.014 on the three-class Wine problem), which is consistent with the sign inversion of the low-frequency, label-carrying eigen-directions predicted by Eq.~\eqref{eq:A0min} for the indefinite operator, rather than with random degradation; reconciling this behaviour with the markedly higher accuracy of the same $\mu=0$ system under the lower supervision budgets of Table~\ref{tab:lowlabel} is part of the per-seed analysis identified in Section~\ref{sec:discussion}. Regularization also reduces split-to-split variability on most benchmarks. Optdigits is a notable exception, exhibiting large standard deviations for the random and quantum-derived diagonals and therefore strong sensitivity to the labelled split. No single diagonal construction dominates uniformly. Random $R$ performs best on Iris and Wine (and jointly best on Breast Cancer), while the quantum-derived variants are competitive on Breast Cancer, Heart Disease, Optdigits, and Cora. The degree-based diagonal is the exception among the nonzero variants: it matches the others on the four small tabular datasets but collapses on German Credit (0.367), Optdigits (0.272), and Cora (0.128), where it is no better than, or worse than, the unregularized arm. This is consistent with the mechanism of Proposition~\ref{prop:pd}: normalization by $\max_j d_j$ drives the entries $d_i/\max_j d_j$ of low-degree nodes toward zero (on Cora, $d_{\min}=1$), so that $\mu\min_i r_i\ge\alpha_2$ fails and the operator is not repaired; we have not verified this explanation per seed. The trained and untrained $\Rt$ rows are identical at the displayed precision on five of the seven datasets, including their standard deviations, and differ only on Iris and Cora, where training slightly lowers the mean. SPSA optimization therefore contributes nothing measurable beyond the initial quantum-compatible encoding under the present protocol.
Overall, the results attribute the main performance gains to diagonal stabilization rather than to the learned structure of $\Rt$, while the large Optdigits variability and the dataset-dependent ranking of the regularizers preclude a universal conclusion.

\begin{table}[t]
\centering
\caption{Ablation of the diagonal regularization matrix in the Poisson system. Results are mean accuracy $\pm$ standard deviation over $N_{\mathrm{seed}}=20$ paired stratified splits at the 30\%-labeled reference setting, computed only on unlabeled nodes. All variants share identical graph constructions, data splits, source terms, and reference hyperparameters; the no-regularization variant differs only by setting $\mu=0$. The highest mean accuracy per dataset at the displayed precision is shown in bold; ties are bolded jointly.}
\label{tab:ablation}
\resizebox{\textwidth}{!}{%
\begin{tabular}{@{}l c c c c c c c@{}}
\toprule
\textbf{Variant} & \textbf{Iris} & \textbf{Wine} & \textbf{Breast Canc.} & \textbf{Heart Dis.} & \textbf{German Cr.} & \textbf{Optdigits} & \textbf{Cora}\\
\midrule
No regularization ($\mu=0$) & 0.094\pmm0.058 & 0.014\pmm0.008 & 0.338\pmm0.049 & 0.343\pmm0.058 & \B{0.637}\pmm0.023 & 0.123\pmm0.026 & 0.152\pmm0.030\\
Random $R$ & \B{0.940}\pmm0.023 & \B{0.964}\pmm0.016 & \B{0.955}\pmm0.008 & 0.821\pmm0.019 & 0.619\pmm0.013 & 0.624\pmm0.440 & 0.840\pmm0.009\\
Degree-based $R$ & 0.927\pmm0.033 & 0.961\pmm0.018 & 0.951\pmm0.008 & 0.821\pmm0.017 & 0.367\pmm0.036 & 0.272\pmm0.145 & 0.128\pmm0.020\\
Untrained $\Rt$ & 0.936\pmm0.025 & 0.956\pmm0.020 & \B{0.955}\pmm0.007 & \B{0.822}\pmm0.016 & 0.623\pmm0.011 & \B{0.730}\pmm0.381 & \B{0.841}\pmm0.008\\
Trained $\Rt$ (IPQSSL) & 0.934\pmm0.025 & 0.956\pmm0.020 & \B{0.955}\pmm0.007 & \B{0.822}\pmm0.016 & 0.623\pmm0.011 & \B{0.730}\pmm0.381 & 0.831\pmm0.014\\
\bottomrule
\end{tabular}}
\end{table}

\section{Circuit-Level Diagnostics}
\label{sec:circuit}

The predictive results evaluate the complete hybrid pipeline, whereas the circuit-level diagnostics analyse the quantum-compatible representation itself. We first define normalized entanglement entropy and a randomized-benchmarking-inspired fidelity proxy, then interpret these quantities jointly before examining their behaviour under depth and width scaling. The diagnostics are computed on four of the seven benchmarks (Iris, Wine, Heart Disease, German Credit); Breast Cancer, Optdigits, and Cora are omitted because the per-node statevector simulations required by the full depth and width sweeps become computationally prohibitive at their larger sample sizes ($n=569$, $n=5620$, and $n=2485$, respectively).

They are not used as substitutes for classification accuracy and do not establish quantum advantage; rather, they quantify how the circuit architecture trades representational structure against noise sensitivity in the QR-based encoding mechanism.

\subsection{Normalized Entanglement Entropy}
\label{sec:entanglement}

This subsection quantifies the amount of bipartite entanglement generated by the variational circuit, providing an architecture-level measure of the internal quantum correlations independently of classification performance. For the encoded state after application of the variational circuit,
\begin{equation}
\ket{\phi_i(\theta)} = V(\theta)\ket{\psi_i},
\end{equation}
let $A$ denote a predefined subsystem and $\bar{A}$ its complement. The corresponding reduced density operator is
\begin{equation}
\rho_A = \Tr_{\bar{A}}\bigl(\ket{\phi_i(\theta)}\bra{\phi_i(\theta)}\bigr).
\end{equation}
The bipartite von Neumann entropy is
\begin{equation}
S_A = -\Tr\bigl(\rho_A\log_2\rho_A\bigr).
\end{equation}
Let
\begin{equation}
d_A = \dim(\mathcal{H}_A).
\end{equation}
The normalized entanglement entropy is defined as
\begin{equation}
\SA = \frac{S_A}{\log_2 d_A}.
\end{equation}
Consequently,
\begin{equation}
0\le\SA\le 1 .
\end{equation}

For every circuit width $q$, subsystem $A$ comprises the first $\lceil q/2\rceil$ qubits in the register ordering of Section~\ref{sec:qr}, so that $d_A=2^{\lceil q/2\rceil}$; the complement $\bar{A}$ contains the remaining qubits. This balanced bipartition is fixed prior to evaluation and is identical across datasets and configurations.

\subsection{Randomized-benchmarking-inspired fidelity}
\label{sec:rb}

We also report a randomized-benchmarking-inspired average fidelity,
\[
\Favg,
\]
as a proxy for gate-level noise in the chosen circuit architecture~\cite{knill2008rb,magesan2011rb,magesan2012rb}. This diagnostic is not a hardware-executed randomized-benchmarking experiment; it is a simulation-based indicator computed under a simplified noise model.

Precisely, $\Favg$ denotes the average fidelity of a single composite circuit layer---one layer comprising the $q$ single-qubit rotation triples and the ring of $q$ CNOT entanglers of Eq.~\eqref{eq:layer}---estimated from the decay of the survival probability of random gate sequences under a depolarizing noise channel applied to every gate, in the spirit of randomized benchmarking. Being a per-layer quantity, $\Favg$ is independent of the depth $L_q$ by construction; it decreases with the width $q$ because each layer then contains more noisy gates. It likewise depends only on the configuration and the noise model, not on the encoded data, and is therefore identical across datasets for fixed $q$; it is reported once per configuration in Tables~\ref{tab:depth} and~\ref{tab:width}.

The relevant trade-off is then the following: at fixed width, the accumulated circuit-level error grows with depth as roughly $\Favg^{L_q}$ even though the per-layer fidelity does not change, while increasing the width lowers the per-layer fidelity itself. Deeper or wider circuits may offer richer representations, but they become less reliable under NISQ constraints.

\subsection{Joint interpretation of diagnostics}

The entanglement and fidelity diagnostics should be interpreted jointly. Increasing circuit width or depth can modify the expressivity of the QR-based representation, but it also increases exposure to noise. Therefore, the circuit architecture should not be selected solely to maximize expressivity. It must also preserve a feasible noise profile.

The present diagnostics support a conservative interpretation of the quantum component. The circuit provides a structured mechanism for generating node-wise diagonal weights, but the diagnostics alone do not prove that these weights are superior to classical diagonal alternatives. Predictive performance and ablation results remain the primary evidence of the method's usefulness.

\subsection{Numerical Results}

Having defined the two circuit diagnostics, we now evaluate how they change under controlled modifications of circuit depth and width. The purpose is to identify whether additional circuit resources produce meaningful representational changes or primarily increase noise-sensitive overhead.

\paragraph{Evaluation of entanglement and Randomized-Benchmarking (RB) with changing layers.}
We evaluate how the quantum diagnostics of the IPQSSL model change as the number of variational layers $L_q$ increases. This analysis examines the balance between circuit expressivity and reliability: deeper circuits may generate stronger entanglement, but they may also accumulate more noise. We therefore report normalized bipartite entanglement entropy $\SA$ and the randomized benchmarking average fidelity $\Favg$. The former measures the relative amount of entanglement produced by the trained circuit, while the latter quantifies circuit reliability under the RB noise model. This evaluation is used as a diagnostic analysis, not as direct evidence of quantum advantage.

\begin{table}[t]
\centering
\caption{Normalized entanglement entropy $\SA$ (mean $\pm$ std over seeds) as a function of circuit depth $L_q$ at fixed width $q=4$. The randomized-benchmarking average fidelity $\Favg$ is a per-layer quantity that depends only on the configuration $(q,L_q)$ and the noise model, not on the encoded data, and is depth-independent by construction (Section~\ref{sec:rb}); it is reported once per depth in the final row. Subsystem $A$ comprises the first $\lceil q/2\rceil$ qubits (Section~\ref{sec:entanglement}).}
\label{tab:depth}
\small
\begin{tabular}{@{}l c c c@{}}
\toprule
& $L_q=2$ & $L_q=6$ & $L_q=10$\\
\midrule
German & 0.6258\pmm0.0095 & 0.6299\pmm0.0403 & 0.6467\pmm0.0151\\
Heart & 0.6427\pmm0.0247 & 0.6091\pmm0.0177 & 0.6066\pmm0.0164\\
Iris & 0.6926\pmm0.0397 & 0.5534\pmm0.0344 & 0.5469\pmm0.0568\\
Wine & 0.6527\pmm0.0094 & 0.5755\pmm0.0438 & 0.5893\pmm0.0731\\
\midrule
RB $\Favg$ & 0.9685\pmm0.0140 & 0.9685\pmm0.0140 & 0.9685\pmm0.0140\\
\bottomrule
\end{tabular}
\end{table}

To disentangle representational gains from accumulated hardware overhead, we vary the number of variational layers $L_q$ at a fixed width of $q=4$ and track two complementary diagnostics: the normalized entanglement entropy $\SA$, which quantifies the internal quantum correlations of the trained circuit, and the randomized-benchmarking average fidelity $\Favg$, which measures gate-level robustness. Interpreting the two jointly distinguishes genuine increases in expressivity from depth that merely adds noise-sensitive gates.

The results show that depth saturates early. On German Credit, Heart Disease, and Wine, $\SA$ stays within a narrow band as $L_q$ grows from 2 to 10 at $q=4$ (for Wine, roughly $[0.58,0.65]$; Table~\ref{tab:depth}), a spread comparable to the per-seed standard deviation. Iris shows the largest depth dependence, decreasing from 0.69 at $L_q=2$ to 0.55 at $L_q=6$, after which it too is flat (0.5469 at $L_q=10$). In all cases the entanglement diagnostic settles by $L_q\simeq 6$, indicating that additional layers primarily add noise-sensitive overhead rather than systematically enriching the representation. Classification accuracy likewise does not vary with depth beyond seed-level variation, so additional layers yield neither richer correlations nor improved generalization. The per-layer fidelity $\Favg$ is depth-independent by construction (Section~\ref{sec:rb}) and equals 0.9685 throughout the sweep, so the accumulated circuit-level error grows with $L_q$ (roughly as $\Favg^{L_q}$); additional layers therefore strictly increase accumulated noise while the entanglement and accuracy diagnostics remain flat. Beyond the saturation point, increasing $L_q$ contributes overhead without commensurate gains in expressivity or accuracy. This controlled-depth sweep motivates the width-scaling study in Section~\ref{sec:width}, where the qubit count---rather than the depth---is shown to be the axis along which the diagnostics actually move.

\begin{figure}[H]
\centering
\includegraphics[scale=.55]{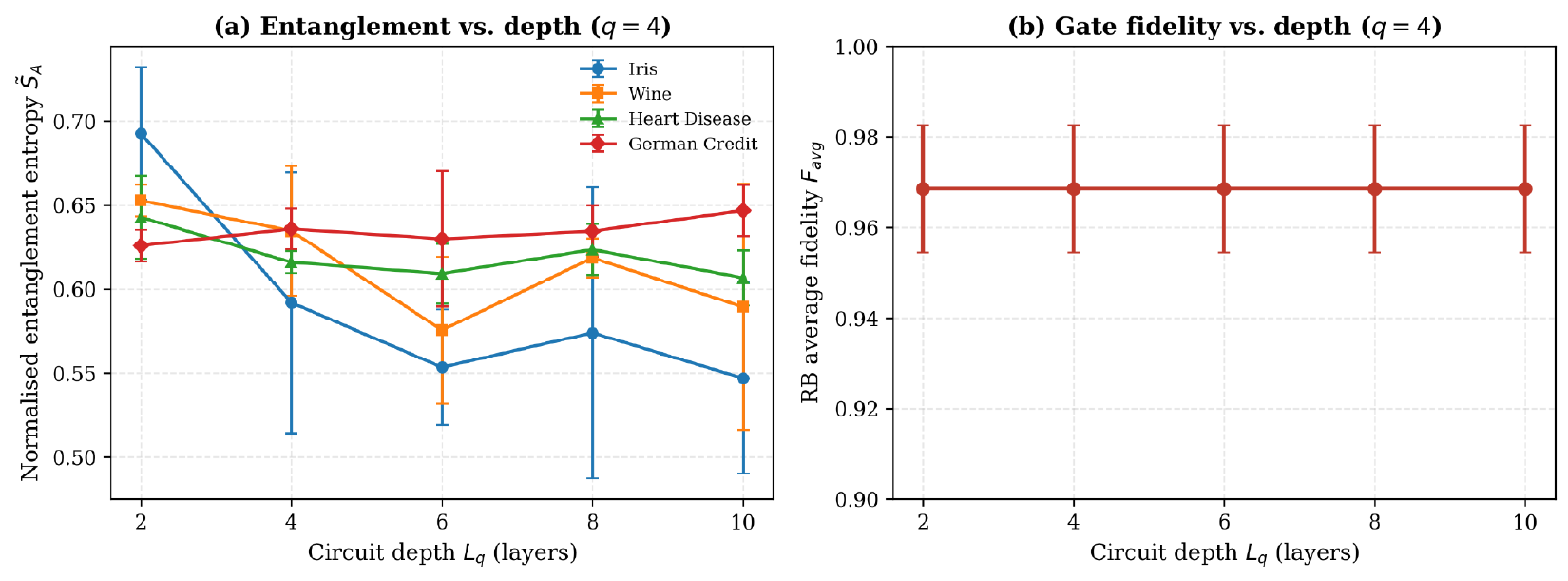}
\caption{Circuit-depth diagnostics at fixed width $q=4$. Panel~(a) reports normalized entanglement entropy as a function of variational depth $L_q$, and Panel~(b) reports the randomized-benchmarking-inspired average fidelity $\Favg$. The results support the interpretation that increasing depth does not provide a systematic diagnostic gain after early saturation.}
\label{fig:depth}
\end{figure}

Under depth scaling at $q=4$ (Figure~\ref{fig:depth}), the normalized entanglement entropy stays in a band around 0.6 and the per-layer RB fidelity is constant at 0.9685 (by construction), so depth is not the effective axis once predictive performance has saturated. By contrast, increasing the qubit count moves the diagnostics substantially: as detailed in Section~\ref{sec:width}, raising $q$ at fixed depth lowers the dataset-mean normalized entanglement (from 0.653 at $q=4$ to 0.442 at $q=10$) while simultaneously reducing the per-layer gate fidelity. Width is therefore the more consequential architectural axis, but it must be traded against gate-level noise rather than treated as a free source of additional quantum structure.

\paragraph{Evaluation of entanglement and RB with changing qubits.}
We evaluate how the quantum diagnostics of IPQSSL change as the number of qubits $q$ increases. This analysis studies the trade-off between representational capacity and circuit reliability: more qubits enlarge the Hilbert space and may support richer quantum correlations, but they can also increase circuit complexity and noise sensitivity~\cite{mcclean2018barren,cerezo2021cost,holmes2022expressibility}. We therefore report the normalized bipartite entanglement entropy $\SA$ and randomized-benchmarking average fidelity $\Favg$. These diagnostics assess whether increasing $q$ produces meaningful entanglement while preserving stable circuit fidelity.

\begin{table}[t]
\centering
\caption{Normalized entanglement entropy $\SA$ (mean $\pm$ std over seeds) as a function of circuit width $q$ at fixed depth $L_q=2$; the $q=4$ column repeats the corresponding column of Table~\ref{tab:depth} for ease of comparison. The data-independent per-layer RB average fidelity $\Favg$ is reported once per width in the final row; it decreases with $q$ because each composite layer contains more noisy gates (Section~\ref{sec:rb}). Subsystem $A$ comprises the first $\lceil q/2\rceil$ qubits (Section~\ref{sec:entanglement}).}
\label{tab:width}
\small
\begin{tabular}{@{}l c c c c@{}}
\toprule
& $q=4$ & $q=6$ & $q=8$ & $q=10$\\
\midrule
German & 0.6258\pmm0.0095 & 0.5639\pmm0.0324 & 0.5371\pmm0.0688 & 0.4242\pmm0.0070\\
Heart & 0.6427\pmm0.0247 & 0.5824\pmm0.0248 & 0.5231\pmm0.0318 & 0.4534\pmm0.0516\\
Iris & 0.6926\pmm0.0397 & 0.5922\pmm0.0565 & 0.5281\pmm0.0287 & 0.4582\pmm0.0744\\
Wine & 0.6527\pmm0.0094 & 0.5937\pmm0.0473 & 0.5276\pmm0.0223 & 0.4339\pmm0.0375\\
\midrule
RB $\Favg$ & 0.9685\pmm0.0140 & 0.9452\pmm0.0061 & 0.9270\pmm0.0113 & 0.9065\pmm0.0178\\
\bottomrule
\end{tabular}
\end{table}

\subsection{Entanglement Scaling and Gate-Level Fidelity}
\label{sec:width}

\begin{figure}[H]
\centering
\includegraphics[scale=.55]{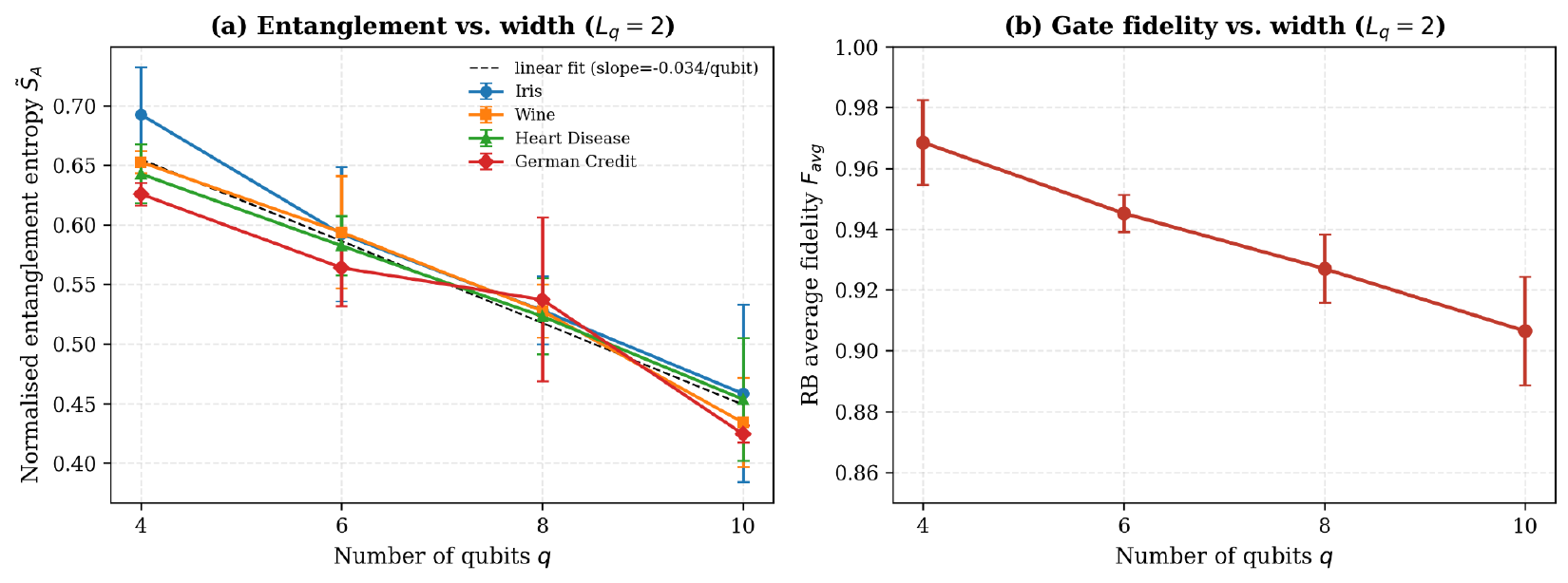}
\caption{Circuit-width diagnostics at fixed depth $L_q=2$. Increasing the number of qubits reduces the normalized entanglement entropy and lowers the randomized-benchmarking-inspired fidelity, illustrating the expressivity--noise trade-off of the QR-based circuit representation.}
\label{fig:width}
\end{figure}

At fixed depth $L_q=2$, we examine how the circuit diagnostics change as the qubit count increases. Table~\ref{tab:width} and Figure~\ref{fig:width} show that raising $q$ from 4 to 10 decreases the dataset-mean normalized entanglement entropy from about 0.65 to 0.44, while the randomized-benchmarking-inspired per-layer fidelity decreases from about 0.97 to 0.91~\cite{knill2008rb,magesan2012rb,cross2019qv}. Thus, increasing width is not a free source of quantum structure: under a fixed-depth ansatz, the generated entanglement becomes diluted across the larger register, while gate-level error accumulates. These quantities are therefore used only as circuit diagnostics, not as evidence of quantum advantage or predictive superiority.

\section{Discussion}
\label{sec:discussion}

The predictive, ablation, spectral, and circuit-level findings combine into a conditional rather than universal claim. Regularized Poisson propagation with a bounded diagonal term improves upon Original Poisson when the similarity graph is well aligned with the data geometry---on Iris and Wine at all but the single-label budget, and at every budget on German Credit, where, however, every method remains below the majority-class rate---but not on Breast Cancer, Heart Disease, Optdigits, or Cora, although the deficit on the two largest graphs narrows as the label budget grows. This dataset- and budget-dependence is itself a finding: diagonal regularization is not a free improvement.

The five-variant ablation attributes the gains primarily to diagonal stabilization itself: a bounded diagonal whose entries stay away from zero---random or circuit-generated---recovers the bulk of the improvement, whereas the degree-based diagonal does so only on the four small tabular datasets and collapses on German Credit, Optdigits, and Cora, where many normalized degrees are close to zero and the sufficient condition of Proposition~\ref{prop:pd} is not met. Supervised SPSA training of the circuit parameters contributes essentially nothing under the present protocol. The spectral analysis supplies the mechanism: by Eq.~\eqref{eq:A0min}, at the default $\alpha_2=0.1$ the unregularized symmetrized operator is indefinite on every dataset ($\lambda_{\min}(\widetilde{A}_0)<0$; Table~\ref{tab:spectral}), with negative and near-zero eigenvalues precisely in the low-frequency, label-carrying directions, so the diagonal term $\mu R$ acts as spectral repair rather than representation learning. Part of what it repairs is a deficit introduced by the default $\alpha_2$ itself; the $\alpha_2=0$, $\mu=0$ control arm required to disentangle sign repair from genuine conditioning improvement is identified as a necessary completion of the ablation.

The contribution of the quantum component is stated precisely: at the register sizes considered, the circuit provides a deterministic, bounded, hardware-compatible inductive bias---not a computational advantage---whose value is already present at initialization. The circuit-generated weights are competitive with, but not demonstrably superior to, simpler diagonal constructions, with the partial exception of the two largest graphs (Optdigits, Cora), where the quantum diagonal (untrained on Cora; trained and untrained jointly on Optdigits) attains the highest mean accuracy among all variants, albeit by margins that are small relative to the split-to-split standard deviation (0.841 versus 0.840 on Cora; 0.730 versus 0.624 with standard deviations near 0.4 on Optdigits). The decisive open comparison is a matched-capacity classical parameterization of the same quadratic form; the quadratic-regularized arm of Table~\ref{tab:lowlabel} is a fixed scalar control rather than such a matched-capacity parameterization and therefore does not close this gap.

Circuit diagnostics favor shallow, narrow circuits: depth saturates early (accuracy and entanglement entropy are flat beyond $L_q\simeq 6$), while widening from $q=4$ to $q=10$ dilutes the normalized entanglement from approximately 0.65 to 0.44 and reduces per-layer gate fidelity from approximately 0.97 to 0.91. Five limitations bound the conclusions: the absence of matched-capacity classical baselines; high split-to-split variance on Optdigits and the below-chance behaviour of the unregularized arm at the 30\% setting, both requiring per-seed analysis; reliance on a single accuracy metric, with German Credit constituting a negative result for all propagation methods considered; classical simulation of all quantum components; and the unquantified share of the ablation gains attributable to undoing the $\alpha_2$-induced spectral shift.

\section{Conclusion}
\label{sec:conclusion}

The conclusion separates the contributions established by the present theory and experiments from the quantum capabilities that remain in the pipeline. We introduced IPQSSL, a quantum-informed regularized Poisson framework for graph-based semi-supervised classification, and evaluated it under a protocol designed to separate the contribution of diagonal stabilization from that of the quantum-generated weights. The contributions that survive this protocol are the following. On the theoretical side, we derived a symmetrized, Hermitian, provably block-encoding-compatible formulation of the regularized Poisson operator, with explicit spectral conditions for positive definiteness and a clean separation between the classical NISQ-oriented implementation and a future fault-tolerant linear-system pathway. On the empirical side, we delineated when diagonal regularization of the Poisson system helps: on Iris and Wine at all but the single-label budget and at every budget on German Credit (where, however, all methods remain below the majority-class rate), but not on Breast Cancer, Heart Disease, Optdigits, or Cora---and we traced the failure of the unregularized system to the indefiniteness of the propagation operator at the standard hyperparameters, so that the regularizer's primary role in this regime is spectral repair. The ablation shows that this repair is achieved by generic bounded diagonals as effectively as by trained quantum-derived ones, and that variational optimization adds nothing under the present protocol; we report this finding as measured, since an honest negative delineation is a prerequisite for credible claims of quantum utility in hybrid graph learning.

Future work follows directly from the limitations. The matched-capacity classical baselines identified in Section~\ref{sec:discussion} are the decisive comparison for the encoding itself, and the positive-definite $\alpha_2=0$ control arm identified in Section~\ref{sec:discussion} is the cheapest experiment that would sharpen the causal attribution of the ablation gains. Per-seed analysis of the Optdigits bimodality and of the below-chance behaviour of the unregularized arm, balanced-accuracy reporting on imbalanced data, and paired significance testing across the full baseline set would complete the empirical protocol. On the quantum side, spectrum-aware observables that target $\lambda_{\min}(\At)$ directly, rather than generic $Z_0$ measurements, would align the circuit output with the mechanism the ablation shows to matter, and hardware execution of the shallow $(q,L_q)=(4,2)$ configuration identified by the diagnostics would replace the simulated fidelity proxy with measured values. The framework's value is that each of these steps is well-posed: the operator to be solved, the assumptions under which it block-encodes, and the component of the pipeline each experiment tests are all stated explicitly.

\paragraph{Author contributions.}
H.G. conceived the framework, designed and implemented the experiments, performed the formal analysis, prepared all visualizations, and drafted the manuscript. F.B. contributed to the mathematical formulation and discussion of schemes and reviewed and edited the manuscript. H.M. contributed to the methodology and reviewed and edited the manuscript. J.N. and M.C. provided supervision, project administration, and revision of the manuscript. All authors approved the final version of the submitted manuscript.

\paragraph{Data availability.}
The tabular datasets used in this study (Iris~\cite{uci_iris}, Wine~\cite{uci_wine}, Breast Cancer (Wisconsin Diagnostic)~\cite{uci_breast}, Heart Disease (Statlog)~\cite{uci_heart}, German Credit~\cite{uci_german}, and Optdigits~\cite{uci_optdigits}) are publicly available from the UCI Machine Learning Repository. The Cora citation network~\cite{sen2008collective} is publicly available in the Planetoid distribution~\cite{yang2016planetoid}.

\paragraph{Code availability.}
The experiment scripts, per-seed result arrays, and figure-generation code supporting this study are available at \url{https://github.com/micss-lab/Quantum_Poisson_SSL}

\paragraph{Acknowledgements.}
The authors thank MICSS Lab and the University of Antwerp for fostering an excellent research environment.
\section*{Funding}
No funding was received for conducting this study.

\section*{Competing interests}
The authors declare that they have no financial or non-financial competing interests.
\bibliographystyle{unsrtnat}
\IfFileExists{Main.bib}{\bibliography{Main}}{}

\end{document}